\documentclass[12pt]{article}
\usepackage{fullpage}
\usepackage[T2A]{fontenc}
\usepackage[koi8-r]{inputenc} 
\usepackage{graphicx} 

\begin{document}
\title{Zipf's Law and Avoidance of Excessive
  Synonymy\footnote{Prepublication draft. Submitted to Cognitive Science.}}
\author{D.~Yu.~Manin, {\tt manin@pobox.com}}
\maketitle
\abstract{Zipf's law states that if words of language are ranked in the order of
  decreasing frequency in texts, the frequency of a word is inversely
  proportional to its rank. It is very robust as an experimental
  observation, but to date it escaped satisfactory theoretical explanation. We
  suggest that Zipf's law may arise from the evolution of word
  semantics dominated by expansion of meanings and competition of synonyms.}
\par

\section*{Introduction} 
Zipf's law may be one of the most enigmatic and controversial
regularities known in linguistics. It has been alternatively billed as
the hallmark of complex systems and dismissed as a mere artifact of data
presentation. Simplicity of its formulation, experimental universality
and robustness starkly contrast with obscurity of its meaning. In its
most straightforward form \cite{Zipf49}, it states that if words of a
language are ranked in the order of decreasing frequency in texts, the
frequency is inversely proportional to the rank,
\begin{equation}
f_k\propto k^{-1} \label{zipf-1}
\end{equation}
where $f_k$ is the frequency of the word with rank $k$. As an example,
Fig.~1 is a log-log plot of frequency vs. rank for a frequency
dictionary of Russian language \cite{FreqDict,Sharoff2002}. The dictionary is
based on a corpus of 40 million words, with special care taken to
prevent data skewing by words with high concentration in particular
texts (like the word {\it hobbit} in a Tolkien sequel). 

\begin{figure}[ht]
\caption{Zipf's law for Russian language}\label{fig:1}
\begin{center}
\input{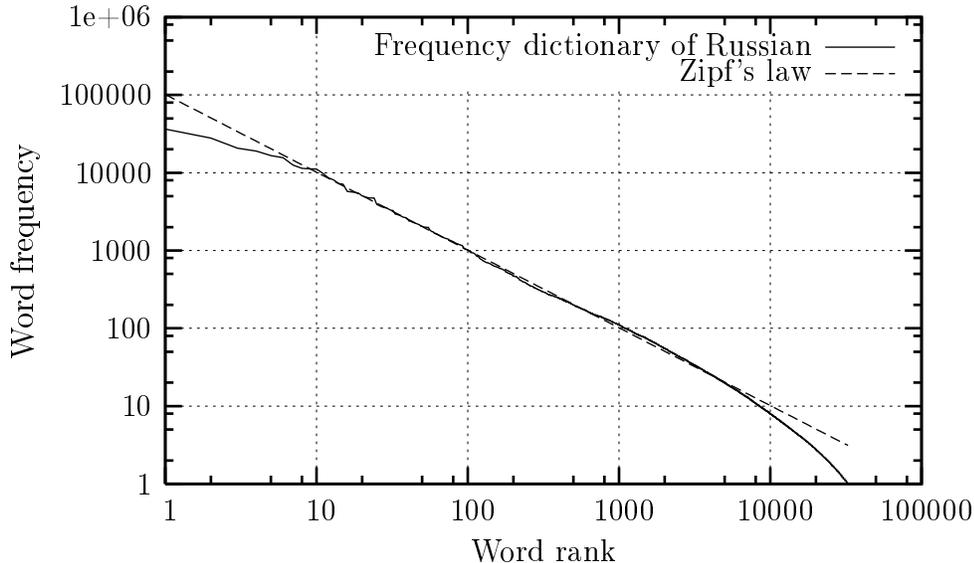}
\end{center}
\end{figure}

Zipf's law is usually presented in a generalized form where the power
law exponent may be different from $-1$, 
\begin{equation}
f_k\propto k^{-B}. \label{zipfB}
\end{equation}
Equivalently, it can be represented as a statement about the
distribution function of words according to their frequency, 
\begin{equation}
P(f)\propto f^{-\beta}, \beta=B+1, \label{zipfbeta}
\end{equation}
where $P(f)df$ represents the fraction of words with frequencies in
$[f, f+df]$.

According to \cite{FerrerVar05}, where an extensive bibliography is
presented, various subsets of the language obey the generalized Zipf's
law (\ref{zipfB}). Thus, while the value of $B\approx 1$ is typical for single
author samples, different values, both greater and less than 1,
characterize speech of schizophrenics and very young children,
military communications, or subsamples consisting of nouns only.

Here we concentrate on the whole language case and do not consider
these variations. Neither do we attempt to generalize our treatment to
include other power law probability distributions, which are
ubiquitous in natural and artificial phenomena of various nature. The
purpose of this work is to demonstrate that the inverse
proportionality (\ref{zipf-1}) can be explained on purely linguistic
grounds. Likewise, we don't pay special attention to the systematic
deviations from the inverse proportionality at the low-rank and high-rank
ends. 

It is not possible to review the vast literature related to the Zipf's
law. However it appears that the bulk of it is devoted to experimental
results and phenomenological models. Models that would aim at
explaining the underlying cause of the power law and predicting the
exponent are not overabundant. We review models of this type in the
first section. In section 2, we discuss the role in the language of
words/meanings having different degrees of generality. In section 3,
we show that Zipf's law can be generated by some particular
arrangements of word meanings over the semantic space. In Section 4,
we discuss the evolution of word meanings and demonstrate that it can
lead to such arrangements. Section 5 is devoted to numerical modeling
of this process. Discussion and prospects for further studies
constitute section 6. In Appendix A, Mandelbrot's optimization model
is considered in detail, and in Appendix B we discuss proportionality
of word frequency to the extent of its meaning.

\section{Some previous models}

\subsubsection*{Statistical models of Mandelbrot and Simon}

The two most well-known models for Zipf's law in the linguistic domain are
due to two prominent figures in the 20th-century science: Beno{\^\i}t
Mandelbrot, of the fractals fame, and Herbert A. Simon, who is listed
among the founding fathers of AI and complex systems theory\footnote{As a historical aside, it is interesting to
mention that Simon and Mandelbrot have exchanged rather spectacularly
sharp criticisms of each other's models in a series of letters in the
journal {\it Information and Control} in 1959--1961.}.

The simplest possible model exhibiting Zipfian distribution is due to
Mandelbrot \cite{Mandelbrot66} and is widely known as {\it random
typing} or {\it intermittent silence} model. It is just a generator of
random character sequences where each symbol of an arbitrary alphabet
has the same constant probability and one of the symbols is
arbitrarily designated as a word-delimiting ``space''. The reason why
``words'' in such a sequence have a power-law frequency distribution
is very simple as noted by Li \cite{Li92}. Indeed, the number of
possible words of a given length is exponential in length (since all
characters are equiprobable), and the probability of any given word is
also exponential in its length. Hence, the dependency of each word's
frequency on its frequency rank is asymptotically given by a power
law. In fact, the characters needn't even be equiprobable for this
result to hold \cite{Li92}. Moreover, a theorem due to Shannon
\cite{Shan48} (Theorem 3 there) suggests that even the condition of
independence between characters can be relaxed and replaced with
ergodicity of the source. 

\begin{figure}[htp]
\caption{Distribution of words by length.}
\centering\input{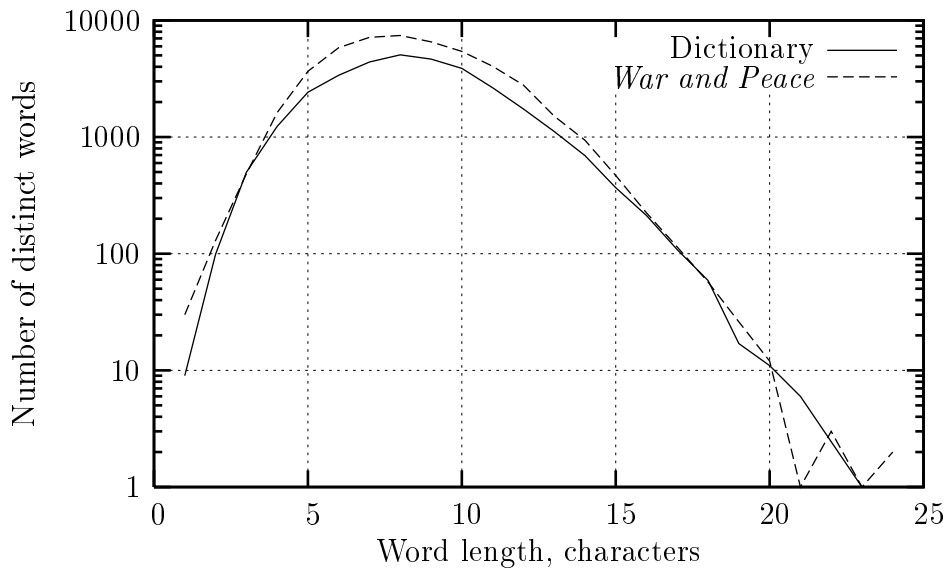}
\label{wordlenDistr}
\end{figure}

Based on this observation, it is commonly held that
Zipf's law is ``linguistically shallow'' (Mandelbrot
\cite{Mandelbrot82}) and does not reveal anything interesting about
the natural language. However it is easy to show that this conclusion
is at least premature. The random typing model itself is undoubtedly
``shallow'', but it cannot be related to the natural language for the
very simple reason that the number of distinct words of the same
length in the real language is far from being exponential in
length. In fact, it is not even monotonic as can be seen in
Fig.~\ref{wordlenDistr}, where this distribution is calculated from a
frequency dictionary of the Russian language \cite{FreqDict} and from
Leo Tolstoy's novel ``War and Peace''. (It also doesn't matter that the
frequency dictionary counts multiple word forms as one word, while
with ``War and Peace'' we counted them as distinct words.)
Thus, even if Zipf's law in natural language is indeed uninteresting,
the random typing model can not prove this. 

Taking a more general view, we observe that Zipf's law is created here
by a simple stochastic process. But human speech is emphatically not a
simple stochastic process. It is a highly structured phenomenon,
driven by extralinguistic needs and stimuli and eventually used for
communication of sentient beings in a real world. If emergence of
Zipf's law may not be surprising in simple models, this doesn't make
it less surprising in such an immensely complex process as speech. Why
should words freely chosen by people to communicate information,
images and emotions, be subject to such a strict probability
distribution?

Another purely statistical model for Zipf's law applicable in various
domains, including language, was proposed by Simon \cite{Simon55},
\cite{Simon57}. It is based on a much earlier work by Yule \cite{Yule1925} who
introduced his model in the context of evolutionary biology
(distribution of species among genera) as early as 1925. Currently,
this and related models are known as {\it preferential attachment} or
{\it cumulative advantage} models, since they describe processes where
the growth rate of an object is proportional to its current
size.

In the linguistic domain, this model in its simplest form describes
writing of a continuous text as a process where the next word
token\footnote{When the same word occurs multiple times in a sequence,
we will speak of {\it word tokens}, {\it occurrences}, or {\it
instances}.} is selected with a constant probability $p$ to be a new,
never before encountered word, and with probability $(1-p)$ to be a
copy of one of the previous word tokens (any one, with equal
probabilities). In this form, the model is not realistic, since it is
well-known that instances of an infrequent word are not distributed
evenly in texts, as the model would predict, but tend to occur in
clusters. However, the model can be significantly relaxed. Namely,
define {\it $n$-word} as a word that has occurred exactly $n$ times in
the preceding text. Suppose that the probability for the next word in
the text to be an (any) $n$-word is equal to the fraction of all
$n$-word tokens in the preceding sequence. Simon showed that this
process still leads to the Zipfian distribution. The model can be
further extended to account for words dropping out of use in such a
way as to preserve the frequency distribution.

In the latter form, Simon's model is compatible with word
clustering. But is it applicable to the natural language?  It is not
quite straightforward to verify the assumptions on which the model is
based. In our calculations using Tolstoy's ``War and Peace'' (about
half a million words in Russian), which we don't report in detail
here, it appears that the assumption of the constant rate of new word
introduction does not hold. Rather, new words are introduced at a rate
that decays approximately as $N^{-0.4}$, where $N$ is the sequence
number of words in text. As for the probability that the next word is
one of $n$-words, it is more or less consistent with the model, except
for the most and the least frequent words. It is not clear
though how critical these departures are for the model.

Simon also argued that the model could be applicable to the language
as a whole, where the birth/death rate describes introduction of
neologisms and words becoming obsolete, while the probability
assumption describes word usage. 

It seems though that the model's expanatory power is not
sufficient. Even if it is correct, we are still left with the question
of {\it why} it is correct. Simon's argument goes approximately as
follows. Suppose the next word choice is described by the probability
$P_{nk} = p_nt_{nk}$, where $P_{nk}$ is the probability that the
$k$-th of the $n$-words will be selected, $p_n$ is the fraction of all
$n$-word tokens in the preceding text, and $t_{nk}$ describes a
``topic'' factor, which favors words appropriate to the topic
currently discussed in the text. It is sufficient to require that
$\sum_k{t_{nk}} = 1$ for all $n$ for the model to work. Thus the model
can even incorporate the idea that people select words according to a
topic rather than randomly. But why would the last equality hold? That
is, why should the selection of some (topical) $n$-words be at the
expense of other $n$-words, and not at the expense of some $m$-words
with $n\ne m$?

More significantly, Simon's model seems to imply that the very fact of
some words being frequent and others infrequent is a pure game of
chance. But in reality, most rare words are rare just because they are
rarely needed. Finally, it is not an
idle question why do we need words with vastly different frequencies
at all. Wouldn't it be more efficient for all words to have about the
same frequency? Simon's model doesn't begin to answer these
questions.

\subsubsection*{Guiraud's semic matrices}

A radically different approach was taken by the French linguist Pierre
Guiraud\footnote{I am grateful to J.D.Apresjan who drew my attention
  to Guiraud's works.}. He suggested that Zipf's law ``would be {\it
  produced} by the structure of the signified, but would be {\it
  reflected} by that of the signifier''
\cite{Guiraud68}. Specifically, suppose that all word meanings can be
represented as superpositions of a small number of elementary
meanings, or {\it semes}. In keeping with the structuralist paradigm,
each seme is a binary opposition, such as {\it animate/inanimate} or
{\it actor/process} (Guiraud's examples). Each seme can be positive,
negative or unmarked in any given word. Assuming that the semes are
orthogonal, so that seme values can be combined with each other
without constraints, with $N$ semes, there can be $2N$ single-seme
words (i.e. words where only one seme is marked), $4N(N-1)$ two-seme
words, and so on. The number of words increases roughly exponentially
with the number of marked semes. On the other hand, assume that 
all semes have the same probability to come up in a speech
situation. Then the probability of a word with $m$ marked semes is
also exponential in $m$. This leads to Zipf's distribution for words.

From the formal point of view, the genesis of Zipf's distribution here
is strikingly similar to that in the random typing model. In both
cases, the number of words and the probability of a word are both
exponential in some parameter (the number of marked semes or the
number of letters respectively). Indeed, by Guiraud's account in
\cite{Guiraud68}, Mandelbrot initially formulated his model in terms
of some hypothetical mental coding units, and only later reformulated
it in terms of letters. In Guiraud's model these coding units turn out
to be the semes. 

This model is very attractive conceptually and heuristically, since it
explains word frequencies as resulting from the language's function as
a vehicle for meaning transfer. However it is too rigid and
schematized to be realistic. It seems very unlikely that the meaning
of any word can be decomposed into an unordered list of about 16
(Guiraud's estimate) binary oppositions, even though theoretically
that would suffice to form enough entries for a typical
dictionary. In addition, the model crucially depends on the assumption
that any combination of semes should be admissible, but even Guiraud's
own examples show that it would be very hard to satisfy this
requirement. Indeed, if {\it actor/process} seme is present with the
value of {\it process}, then {\it animate/inanimate} has to be
unmarked: there are no animate or inanimate verbs. (Some verbs, such
as {\it laugh} imply the animateness of the actor, but that's a
different trait. The point is that there is no verb that would differ
from {\it laugh} only in that it's inanimate -- and that undermines
the notion of unrestricted combinability of semes.) In addition, it
doesn't offer any diachronic perspective.

\subsubsection*{Models based on optimality principles}

Different authors proposed models based on the observation that Zipf's
law maximizes some quantity. If this quantity can be interpreted as a
measure of ``efficiency'' in some sense, then such model can claim
explanatory power. 

Zipf himself surmised in \cite{Zipf49} that this distribution may be a
result of ``effort minimization'' on the part of both speaker and
listener. This argument goes approximately as follows: the
broader\footnote{We will use {\it broad} or {\it generic} on the one
hand and {\it narrow} or {\it specific} on the other to characterize
the {\it extent} or {\it scope} of a word's meaning.} the meaning of a word, the more
common it is, because it is usable in more situations. More common
words are more accessible in memory, so their use minimizes speaker's
effort. On the other hand, they increase the listener's effort,
because they require extra work on disambiguation of diffuse
meanings. As a result of a compromise between speaker and listener, a
distribution emerges.

Zipf did not construct any quantitative model based on these
ideas. The first model of this sort was proposed by Mandelbrot
\cite{Mandelbrot53}. It optimizes the cost of speech production per
bit of information transferred. Let the {\it cost} of producing
word $w_k$ be $C_k$. The word's {\it information content}, or entropy,
is related to its frequency $p_k$ as $H_k = -\log_2{p_k}$. The average
cost per word is given by $C = \sum_k{p_kC_k}$ and the average entropy
per word by $H=-\sum_k{p_k\log_2{p_k}}$. One can now ask what frequency
distribution $\{p_k\}$ satisfying $\sum_k{p_k}=1$ will minimize the
ratio $C/H$. An easy calculation using Lagrange multipliers leads to 
\begin{equation}
p_k = Ae^{-HC_k/C},
\end{equation}
where $A$ is the normalization factor which needs to be chosen so that
all the probabilities sum up to 1. In order to obtain a power law, the
cost $C_k$ needs to be logarithmic in $k$,
$C_k\propto\log{k}$. Mandelbrot derived this formula assuming that the
cost of a word is proportional to its length, and the number of different
words of length $l$ is exponential in $l$. Then, the result becomes
almost trivial, since it's well known that maximum information per
letter is achieved by a random sequence of letters, and we return to
the random typing model. To cite Mandelbrot \cite{Mandelbrot66},
``These variants are fully equivalent mathematically, but they appeal
to [...]  different intuitions [...]''.

As we mentioned above, the assumption that the number of words is
exponential in word length is incorrect (Fig.~\ref{wordlenDistr}). However there is a
different and much more plausible argument for the direct relationship
between cost and rank: $\log_2{k}$ is the number of bits that need to
be specified in order to retrieve the $k$-th word from memory (if
words are stored in the order of decreasing frequency, which is a
natural assumption), and thus a good candidate for a cost estimate.
We leave the detailed treatment of this case for Appendix~A, because it is not
essential for the main argument here.

But once an optimization model is constructed, it is neccessary to
demonstrate that the global optimum can actually be achieved via some
local dynamics which is causal and not teleological. Thus, the famous
principle of least action in mechanics is equivalent to the local
force-driven Newtonian dynamics. In the same way, a soap film on a
wire frame achieves the global minimum of surface area via local
dynamics of infinitesimal surface elements shifting and stretching
under each other's tug. Just like surface elements do not ``know''
anything about the total area of the film, individual words do not
``know'' anything about the average information/cost ratio. 

Interestingly, in the case of Mandelbrot's optimizing model, such a
local dynamics can be proposed. Namely, suppose that if speakers
notice that a word's individual information/cost ratio is below
average (the word has $faded$), they start using it less, and
conversly, if the ratio is favorable, the word's frequency
increases. It turns out that this local dynamics indeed leads to an
establishment of a stable power-law distribution of word frequencies
(see Appendix~A for details).

Even in this form, Mandelbrot's model has two problems. First, the
power law exponent turns out to be very sensitive to the details of
the cost function $C_k$. This lack of robustness is significant,
because the pure logarithmic form of cost function is just a very
rough approximation. The second problem is that the local dynamics
described above as the mechanism for a real language to achieve the
optimum cost ratio, is not realistic. People will not start using a
word like, say, {\it table} more frequently just because it happens to
have a favorable cost ratio. They will use it when they need to refer
to (anything that can be called) a table --- no more, no
less\footnote{To be fair, somehting similar does occur in languages
when so-called expressive synonyms change to regular words. A
well-known example is Russian {\it глаз}, `eye', which initially meant
`pebble', then became expressive for `eye', and gradually
displaced the original word for `eye', {\it око} of Indo-European
descent. Another example is provided by French {\it t\^ete}, `head'
below. But this is a different kind of dynamics involving competition
of two words. It will be considered below.}. And a compelling explanation of Zipf's law has to
comply with this reality.

A different model was proposed by Arapov \& Shrejder
\cite{ArapovShrejder78}. They demonstrated that Zipfian distribution
maximizes a quantity they call {\it dissymmetry}, which is the sum of two
entropies: $\Phi=H+H^*$, where $H$ is the standard entropy that
measures the number of different texts that can be constructed from a
given set of word tokens (some of which are identical), while $H^*$
measures the number of ways {\it the same} text can be constructed
from these tokens by permutations of identical word tokens. The former
quantity is maximized when all word tokens in a text are different,
the latter one when they are all the same, and the Zipfian
distribution with its steep initial decline and long tail provides the
best compromise. This theoretical construct does possess a certain
pleasing symmetry, but its physical meaning is rather obscure, though
the authors claim that $\Phi$ should be maximized in ``complex systems
of natural origin''. 

Balasubrahmanyan and Naranan \cite{BalasubraNaranan2002} take a
similar approach. They too, aim to demonstrate that the language is a
``complex adaptive system'', and that Zipf's law is achieved in the
state of maximum ``complexity''. Their derivation also involves
defining and combining different entropies, some of which are related
to the permutation of identical word tokens in the text. Both
approaches of \cite{ArapovShrejder78} and
\cite{BalasubraNaranan2002}, in our view, have the same two
problems. First, the quantity being optimized is not compellingly
shown to be meaningful. Second, no mechanism is proposed to explain
why and how the language could evolve towards the maximum. To quote
\cite{BalasubraNaranan2002}, 
\begin{quote}
As a general principle, an extremum is the most stable configuration
and systems evolve to reach that state. We do not however understand
the details of the dynamics involved.
\end{quote}

In a recent series of articles by Ferrer i Cancho with coauthors (see
\cite{FerrerLeastEffort05}, \cite{FerrerZipfExponent05} and references
therein) the optimization idea is taken closer to the reality. Ferrer
i Cancho's (hereafter FiC) models significantly differ from the other
models in that they are based on the idea that the purpose of language
is communication, and that it is optimized for the efficiency of
communication. FiC models postulate a finite set of words and a finite
set of objects or stimuli with a many-to-many mapping between the
two. Multiple objects may be linked to the same word because of
polysemy, while multiple words may be linked to the same object
because of synonymy. Both polysemy and synonymy are, indeed, common
features of natural languages. It is assumed that the frequency of a
word is proportional to the number of objects it is linked to. Next,
FiC introduces optimality principles and, in some cases, constraints,
with the meaning of coder's effort, decoder's effort, mutual entropy
between words and objects, entropy of signals, and so on. By
maximizing goal functions constructed from combinations of these
quantities, FiC demonstrated the emergence of Zip's law in phase
transition-like situations with finely tuned parameters.

The treatment in the present work, although quite different in spirit,
shares two basic principles with FiC's models and, in a way, with
Guiraud's ideas. First, we also consider it essential that language is
used for communication and adopt the mapping metaphor of meaning
(although at the early stages of language evolution, control of
behavior rather than communication may have been its primary function
--- see e.g. \cite{ManinGlotto}). Second, we postulate that word
frequency is proportional to the extent, broadness, or generality of
its meaning (see below for a more detailed discussion). But we also
differ from FiC and Zipf in a couple of important aspects. We do not
assume any optimality principles and neither do we use the notion of
least effort. Instead, we show that Zip's law can be obtained as a
consequence of a purely linguistic notion of avoidance of excessive
synonymy. It should be noted that our approach need not be mutually
exclusive with that of FiC. In fact, they may turn out to be
complementary. It may also be compatible with (but providing a deeper
explanation than) Simon's model.

If one is to claim that word frequency in texts is related to some
properties of its meaning, a theory of meaning must be presented
upfront. Fortunately, it doesn't have to be comprehensive, rather
we'll outline a {\it minimal} theory that only deals with the single
aspect of meaning that we are concerned with here: its extent. 

\section{Synonymy, polysemy, semantic space}

The nature of meaning has long been the subject of profound
philosophical discourse. What meaning is and how meanings are
connected to words and statements is not at all a settled
question. But whatever meaning is, we can operate the notion of ``the
set of all meanings'', or ``semantic space'', because this doesn't
introduce any significant assumptions about the nature of meaning
(except, maybe, its relative stability). Of course, we should exercise
extreme caution to avoid assuming any structure on this set which we
don't absolutely need. For example, it would be unwise to think of
semantic space as a Euclidean space with a certain dimensionality
(as is the case with Guiraud's semic matrices). One
could justify the assumption of a metric on semantic space, because we
commonly talk about meanings being more or less close to each other,
effectively assigning a distance to a pair of meanings. However as we
won't need it for the purposes of this work, metric will not be
assumed.

In fact, the only additional structure that we do assume on semantic
space $S$, is a measure. Mathematically, measure on $S$ assigns a
non-negative ``volume'' to subsets of $S$, such that the volume of a
union of two disjoint subsets is the sum of their
volumes\footnote{Many subtleties are omitted here, such as the fact
that a measurable set may have non-measurable subsets.}. We need
measure so that we can speak of words being more ``specific'' or
``generic'' in their meanings. If a word $w$ has a meaning
$m(w)\subset S$, then ``degree of generality'', or ``extent'', or
``broadness'' of its meaning is the measure $\mu(m(w))$, i.e.~the
amount of ground that the word covers in semantic space.\footnote{We
assume that meanings of words correspond to subsets of $S$. It may
seem natural to model them instead with fuzzy subsets of $S$, or,
which is the same, with probability distributions on $S$. However the
author feels that there is already enough fuzziness in this treatment,
so we won't develop this possibility. Meanings may also be considered
as prototypes, i.e.~attractors in semantic space, but our model can be
adapted to this view as well.}  Note that measure does not imply
metric: thus, there is a natural measure on the unordered set of
letters of Latin alphabet (``volume'' of a subset is the number of
letters in it), but to define metric, i.e.~to be able to say that the
distance between {\it a} and {\it b} is, say, 1, we need to somehow
order the letters.

We understand ``meaning'' in a very broad sense of the word. We are
willing to say that any word has meaning. Even words like {\it the}
and {\it and} have meanings: that of definiteness and that of
combining respectively. We also want to be able to say that such words
as {\it together}, {\it joint}, {\it couple}, {\it fastener} have
meanings that are subsets of the meaning of {\it and}. By that we mean
that in any situation where {\it joint} comes up, {\it and} also comes
up, though maybe implicitly (whatever that means). We do not make
distinction between connotation and denotation, intension and
extension, etc. This means that ``semantic space'' $S$ may include
elements of very different nature, such as the notion of a mammal, the
emotion of love, the feeling of warmth, and your cat Fluffy. Such
eclecticism shouldn't be a reason for concern, since words are in fact
used to express and refer to all these kinds of entities and many
more.

We only deal with isolated words here, without getting into how the
meaning of {\it this dog} results from the meanings of {\it this} and
of {\it dog}. Whether it is simply a set theoretic intersection of
{\it thisness} and {\it dogness} or something more complicated, we
don't venture to theorize. The biggest problem here is probably that
the semantic space itself is not static, new meanings are created all
the time as a result of human innovation in the world of objects, as
well as in the world of ideas: poets and mathematicians are especially
indefatigable producers of new meanings\footnote{For a much deeper discussion
see \cite{ManinRKS}. In particular, it turns out that the rich
paraphrasing capacity of language may paradoxically be an evidence of high
referential efficiency.}. However, when dealing with individual words, as
is the case with Zipf's law, one can ignore this instability, since
words and their meanings are much more conservative, and only a small
fraction of new meanings created by the alchemy of poetry and
mathematics eventually claim words for themselves.

Note that up to now we didn't have to introduce any structure on $S$,
not even measure. Even the cardinality of $S$ is not specified, it
could be finite, countable or continuous. But we do need measure for
the next step, when we assume that the frequency of the word $w$ is
proportional to the extent of its meaning, i.e.~to the measure 
$\mu(m(w))$. The more generic the meaning, the more frequent the word, and
vice versa, the more specific the meaning, the less frequent the word.

We don't have data to directly support this assumption, mostly because we don't
know how to independently measure the extent of a word's meaning. One
could think of ways to do this, such as the length of the word's
dictionary definition or the number of all hyponyms of the given word
(for instance, using
WordNet\footnote{http://wordnet.princeton.edu/}). It would be
interesting to see if word frequency is correlated to such measures,
but we are not aware of any research of this kind. The assumption
itself however appears to be rather natural, and in Appendix~B we
provide some experimental evidence to support it.

It is essential for this hypothesis that we do not reduce meaning to
denotation, but include connotation, stylistical characteristics,
etc. It is easy to see that the word frequency can't be proportional
to the extent of its denotation alone: the word {\it dog} is more
frequent that words {\it mammal} and {\it quadruped}, though its
denotation (excluding figurative senses though) is a strict subset and
thus more narrow\footnote{I owe this example to Tom Wasow.}. But the
frequency of the word {\it mammal} is severely limited by its being a
scientific term, i.e. its meaning extent is wider along the denotation
axis, but narrower along the stylistic axis (``along the axis'' should
be understood metaphorically here, rather than technically). In the
realm of scientific literature, where the stylistic difference is
neutralized, {\it mammal} is quite probably more frequent than {\it
dog}.

It's interesting to note in this connection that according to the
frequency dictionary \cite{FreqDict}, the word {\it собака} `dog' is
more frequent in Russian than even the words {\it животное} and {\it
зверь} `animal, beast', although there is no significant stylistical
differences between them. To explain this, note that of all animals,
only the dog and the horse are so privileged. A possible reason is
that the connotation of {\it animal} in the common language includes
not so much the opposition `animal as non-plant' as the opposition
`animal as non-human'. But the dog and the horse are
characteristically viewed as ``almost-human'' companions, and thus in
a sense do not belong to animals at all, which is why the
corresponding words do not have to be less frequent.

The vocabulary of a natural language is structured so that there are
words of different specificity/generality. According to WordNet, a
rose is a shrub is a plant is an organism is an object is an
entity. There are at least two pretty obvious reasons for this. First,
in some cases we need to refer to any object of a large class, as in
{\it take a seat}, while in other cases we need a reference to a
narrow class, as in {\it you're sitting on a Chippendale}. In the
dialogue (\ref{candy}) two words, the generic one and the specific
one, point to the same object. 
\begin{equation}\label{candy}
\vcenter{
 \hbox{--- I want some Tweakles!}
 \hbox{--- Candy is bad for your teeth.} 
}
\end{equation}

Second, when context provides
disambiguation, we tend to use generic words instead of specific
ones. Thus, inhabitants of a large city environs say {\it I'm going to
the city} and avoid naming it by name. Musicians playing winds call
their instrument a {\it horn}, whether it's a trumpet or a tuba.  Pet
owners say {\it feed the cat}, although the cat has a name, and some
of them perform a second generalization to {\it feed the beast} (also
heard in Russian as {\it накорми животное}). In fact, the word {\it
candy} in the {\it Tweakles} example fulfills both roles at once: it
generalizes to all candies, because all of them are bad for your
teeth, but also it refers to this specific candy by contextual
disambiguation. We even use the ultimate generic placeholders like
{\it thingy} when we dropped it and need somebody to pick it up for
us\footnote{As Ray Bradbury wrote in his 1943 story {\it Doodad}:
''Therefore, we have the birth of incorrect semantic labels that can
be used to describe anything from a hen's nest to a motor-beetle
crankcase. A doohingey can be the name of a scrub mop or a
toupee. It's a term used freely by everybody in a certain culture. A
doohingey isn't just one thing. It's a thousand things.'' WordNet
lists several English words under the definition ``something whose
name is either forgotten or not known''.  Interestingly, some of these
words ({\it gizmo, gadget, widget}) developed a second sense, ``a
device that is very useful for a particular job'', and one ({\it
gimmick}) similarly came to also mean ``any clever (deceptive) maneuver''.}. A
colorful feature of Russian vernacular is the common use of
desemantized expletives as generic placeholders, where whole sentences
complete with adjectives and verbs can be formed without a significant
word. What may not be generally appreciated is that this strategy may, at least in some cases, turn
out to be highly efficient. According to the author V.~Konetsky
\cite{KonetskySL}, radio communications of Russian
WWII fighter pilots in a dogfight environment, where a split-second
delay can be fatal, consisted almost entirely of such pseudo-obscene
placeholder words, as evidenced by recordings. It hardly could have
been so, were it not efficient.

The reason for this tendency to generalize is very probably the
Zipfian minimization of effort for the speaker. A so-called {\it word
frequency effect} is known in psycholinguistics, whereby the more
frequent the word the more readily it is retrieved from memory
(cf. \cite{AkmajianEtAl95}, \cite{Carrol94}). However, contrary to
Zipf, it doesn't seem plausible that such generalization makes
understanding more difficult for the listener. The whole idea of
pitching the speaker against the listener in the effort minimization
tug-of-war appears to fly in the face of communication as an
essentially cooperative phenomenon, where a loss or gain of one party
is a loss or gain of both. Again, we don't have hard data, but
intuitively it seems that when there is only one city in the context
of the conversation, it is even easier for the listener if it's
referred to as {\it the city} rather than {\it Moscow} or {\it New
York}. {\it I'm going to the city} means {\it I'm going you know
where} while {\it I'm going to London} means {\it I'm going to this
one of a thousand places where I could possibly go}. The first
expression is easier not only for the speaker, but for the listener as
well, because one doesn't have to pull out one's mental map of the
world, as with the second expression. Or, put in information theoretic
terms, {\it the city} carries much less information than {\it Shanghai}
because the generic word implies a universal set consisting of one
element, while the proper name implies a much larger universal set of
dozens of toponyms, --- but most of this extra information is junk and has to
be filtered out by the listener, if Shanghai is in fact The City; and
this filtering is a wasted effort.

\section{Zipf's law and Zipfian coverings}

Organization of words over semantic space in such a way that each
element is covered by a hierarchy of words with different extent of
meaning makes a lot of sense. In this way, the speaker can select a
word that refers to the desired element with the desired degree of
precision. Or, rather, the most imprecise word that still allows
disambiguation in the given context. The benefit here is that less
precise words are more frequent, and thus more accessible for both the
speaker and the listener, which can be said to minimize the effort for
both. Another benefit is that such organization is conductive to
building hierarchical classifications, which people are rather
disposed to do (whether that's because world itself is hierarchically
organized, is immaterial here). There are probably other benefits as
well.

\begin{figure}[htp]
\centering
\caption{Example of a hierarchical organization of semantic space.}\label{fig:2}
%\fig{img/fig2}
\includegraphics[width=12cm]{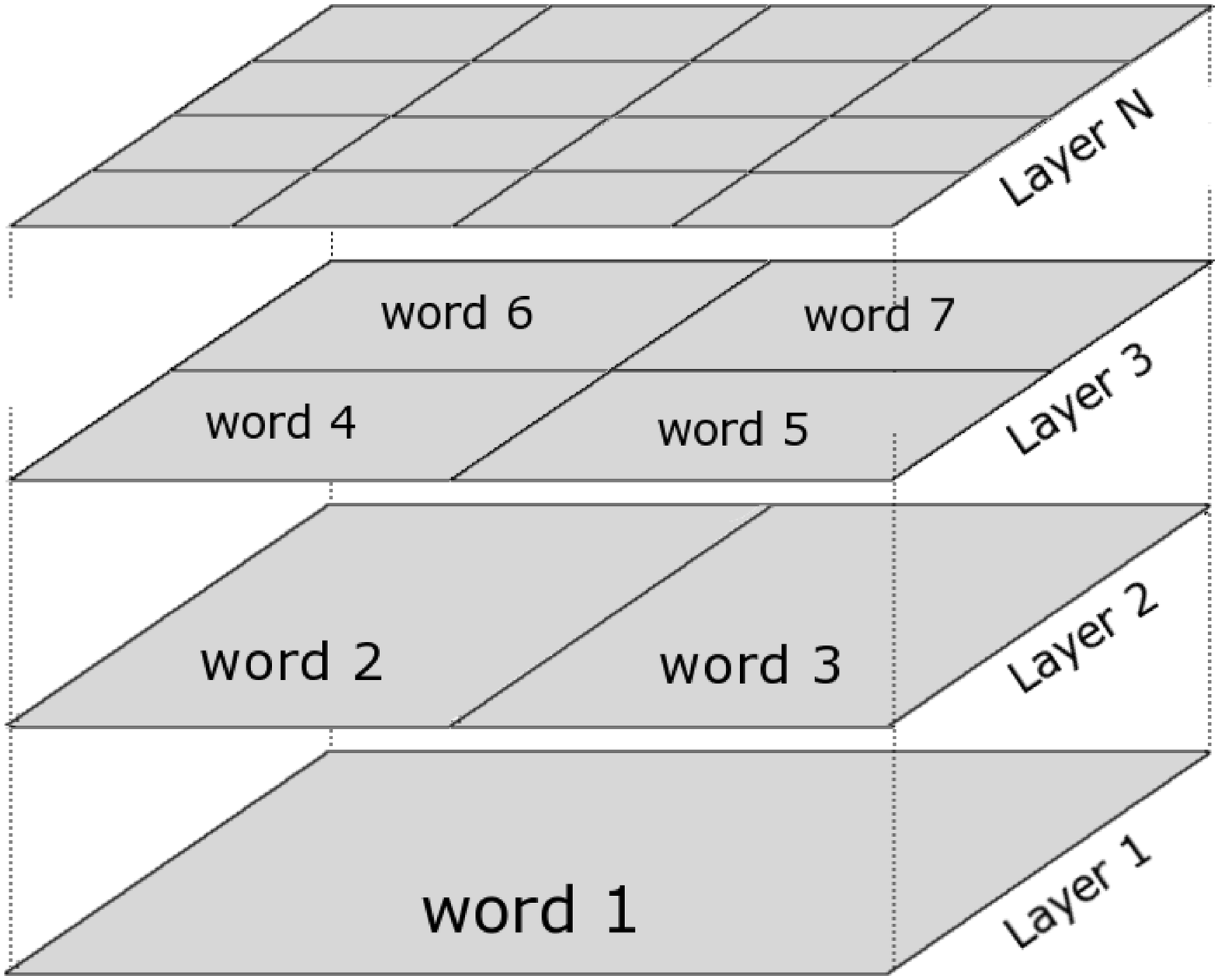}
\caption{Frequency distribution for hierarchical model Fig.~\ref{fig:2}.}\label{fig:3}
\input{img/fig3.tex}
\end{figure}

Here is the simplest possible way to map words to semantic space in
this hierarchical manner: let word number 1 cover the whole of $S$,
words number 2 and 3 cover one-half of $S$ each, words 4 through 7
cover one-quarter of $S$ each, etc. (see Fig.~\ref{fig:2}). It is easy to see
that this immediately leads to Zipf's distribution. Indeed, the extent
of the $k$-th word is
\begin{equation}
\mu_k = 2^{-\lceil \log_2 k\rceil} \label{zipfianmodel}
\end{equation}
Under the assumption that the frequency of a word $f_k$ is proportional to
the extent of its meaning $\mu_k$, this is equivalent to (\ref{zipf-1}),
except for the piecewise-constant character of (\ref{zipfianmodel}),
see Fig.~\ref{fig:3}. What matters here is the overall trend, not the fine
detail. 

Of course, real word meanings do not follow this neat, orderly model
literally. But it gives us an idea of what Zipf's distribution
(\ref{zipf-1}) can be good for. Consider a subset of all words whose
frequency rank is in the range $[k, k\rho]$ with some $k$ and
$\rho>1$. Zipf's distribution has the following property: the sum of
frequencies of words in any such subset depends only
on the scaling exponent $\rho$ (asymptotically with
$k\rightarrow\infty$), since by Riemann's formula, it is bounded by
inequalities
\begin{equation}
\ln\frac{n}{k} = \int_k^n\frac{{\rm d}x}{x} <  \sum_{j=k}^n\frac{1}{j}
< \int_{k-1}^{n-1}\frac{{\rm d}x}{x} = \ln\frac{n-1}{k-1}
\end{equation}
By our basic assumption, word frequency is proportional to the extent of
its meaning. Thus, we can choose $\rho$ so that the words in any
subset $[k, k\rho]$ together could cover the whole semantic space
$S$ without gaps and overlaps: the sum of their meanings' measures
will be equal to the total measure of $S$. Of course, this does not
guarantee that they {\it do} cover $S$ in such a way, but only for Zipf's
distribution such a possibility exists. 

Let us introduce some notation at this point, to avoid bulky
descriptions. Let $S$ be a measurable set with a finite
measure $\mu$. Define {\it covering} of $S$ as an arbitrary sequence of
subsets $C=\{m_i\}, m_i\subset S, \mu(m_i)\ge \mu(m_{i+1})$. Let the {\it
  gap} of $C$ be the measure of the part of $S$ not covered by $C$,
\begin{equation}
{\rm gap}(C) = \mu(S)-\mu(\bigcup m_i),
\end{equation} 
and let {\it overlap} of $C$ be the measure
of the part covered by more than one $m_i$,
\begin{equation}
{\rm overlap}(C) = \mu(\left\{x|x\in {\rm more\ than\ one\ }
m_i\right\}) = \mu\big(\big\{x|x\in\bigcup_{i\ne j}{m_i\cap m_j}\big\}\big)
\end{equation}
 Finally, define {\it
  $(\rho$, k)-layer} of $C$ as subsequence $\{m_i\}, i\in[k, k\rho]$
for any starting rank $k>0$ and some scaling exponent $\rho>1$. 

With these definitions, define {\it Zipfian covering} as an infinite
covering such that for some $\rho$, both gap and overlap of $(\rho,
k)$-layers vanish as $k\rightarrow\infty$. This means that
all words with ranks in any range $[k, k\rho]$ cover the totality of $S$
and do not overlap (asymptotically in $k\rightarrow\infty$). Or, to
look at it from a different point of view, each point in $S$ is
covered by a sequence of words with more and more precise (narrow, specific)
meanings, with precision growing in geometric progression with
exponent $\rho$. Again, this organization of semantic space would make
a lot of sense, since it ensures the homogeneity of the ``universal
classification'': precision of terms increases by a constant factor
each time you descend to the next level. This is why the exponent $B=1$ in
(\ref{zipfB}) is special: with other exponents one doesn't get the
scale-free covering.

The covering in Fig.~\ref{fig:2} is an example of Zipfian covering,
though a somewhat degenerate one. We will not discuss the existence of
other Zipfian coverings in the strict mathematical sense, since the
real language has only a finite number of words anyway, so the limit
of an infinite word rank is unphysical. We need this as a {\it strict}
definition of an idealized model which is presumably in an {\it
approximate} correspondence with reality. 

Note though that since $\sum_1^n{1/j}$ grows indefinitely as
$n\rightarrow\infty$, Zipf's law can be normalized only if cut off at
some rank $N$. The nature of this cut-off becomes very clear in the
present model: the language does not need words with arbitrary
narrow meanings, because such meanings are more efficiently
represented by combinations of words.

However, as noted above, demonstrating that Zipf's law satisfies some
kind of optimality condition alone is not sufficient. One needs to
demonstrate the existence of a plausible local dynamics that could be
responsible for the evolution towards the optimal state. To this end,
we now turn to the mechanisms and regularities of word meaning change.

\section{Zipfian coverings and avoidance of excessive synonymy}

Word meanings change as languages evolve. This is a rule, rather than
an exception (see, e.g. \cite{HockJoseph}, \cite{Maslov}; most of the
examples below come from these two sources). There are various
reasons for semantic change, among them need, other changes in the
language, social factors, ``bleaching'' of old words, etc. Some
regularities can be observed in the direction of the change. Thus, in
many languages, words that denote grasping of physical objects with
hands develop the secondary meaning of understanding, ``grasping of ideas with
mind'': Eng.\ {\it comprehend} and {\it grasp}, Fr. {\it comprendre},
Rus. {\it понимать} and {\it схватывать}, Germ.\ {\it fassen}
illustrate various stages of this development. Likewise, Eng.\ {\it
clear} and Rus.\ {\it ясный}, {\it прозрачный} illustrate the drift
from optical properties to mental qualities. As a less spectacular,
but ubiquitous example consider metonymic extension from action to its
result, as in Eng.\ {\it wiring} and Rus.\ {\it проводка} (idem). There
may also be deeper and more pervasive regularities
\cite{Traugott}. Paths from old to new meanings are usually
classified in terms of metaphor, metonymy, specialization, ellipsis,
etc.\ \cite{CambridgeEncLang}.

Polysemy, multiplicity of meanings, is pervasive in language: ``cases
of monosemy are not very typical'' \cite{Maslov}; ``We know of no
evidence that language evolution has made languages less ambiguous''
\cite{WasowAmbiguity}; ``word polysemy does not prevent people from
understanding each other'' \cite{Maslov}. There is no clear-cut
distinction between polysemy and homonymy, but since Zipf's law deals
with typographic words, we do not have to make this distinction. In
the ``meaning as mapping'' paradigm, one can speak of different {\it
senses}\footnote{This should not be confused with the dichotomy of
{\it sense} and {\it meaning}. Here we use the word {\it sense} as in
``the dictionary gave several senses of the word''.} of a polysemous
word as subsets of its entire meaning. Senses may be separate (cf. {\it
sweet}: `tasting like sugar' and `amiable'
\footnote{Definitions here and below are from 1913 edition of Webster's dictionary.}), they may overlap ({\it
ground}: `region, territory, country' and `land, estate, possession'),
or one may be a strict subset of the other ({\it ball}: `any round or
roundish body' and `a spherical body used to play with').

Note that causes, regularity and paths of semantic change are not
important for our purposes, since we are only concerned here with the
extent, or scope, of meaning. And that can change by three more or
less distinct processes: extension, formation, and disappearance of
senses (although the distinction between extension and
formation is as fuzzy as the distinction between polysemy and
homonymy).

Extension is illustrated by the history of Eng.\ {\it bread} which
  initially meant `(bread) crumb, morsel' (\cite{HockJoseph}, p.~11),
  or Rus.\ {\it палец}, `finger, toe', initially `thumb'
  (\cite{Maslov}, p.~197--198). With extension, the scope of
  meaning increases.

Formation of new senses may cause increase in meaning scope or no
change, if the new sense is a strict subset of the existing ones. This
often happens through ellipsis, such as with Eng.\ {\it car},
`automobile' < {\it motor car} \cite{HockJoseph}, p.~299 or parallel
Rus.\ {\it машина} < {\it автомашина}. In this case, the word initialy
denotes a large class of objects, while a noun phrase or a compound
with this word denotes a subclass. If the subclass is important
enough, the specifier of the phrase can be dropped (via generalization
discussed above), and this elliptic usage is reinterpreted as a new,
specialized meaning.

Meanings can decrease in scope as a result of a sense dropping
out of use. Consider Eng.\ {\it loaf} < OE {\it hlaf},
`bread'. Schematically one can say that the broad sense `bread
in all its forms' disappears, while the more special sense `a
lump of bread as it comes from the oven' persists. Likewise, Fr.\ {\it
chef}, initially `head as part of body', must have first acquired the
new sense `chief, senior' by metaphor, and only then lost the
original meaning.

In the mapping paradigm, fading of archaic words can also be
interpreted as narrowing of meaning. Consider Rus.\ {\it перст},
'finger (arch., poet.)'. The reference domain of this word is almost
the same as that of {\it палец} 'finger (neut.)' (excluding the
sense `toe'), but its use is severely limited because of a
strong flavor. Thus, meaning scope is reduced here along the
connotation dimension. But since we consider both denotation and
connotation as constituents of meaning, narrowing of either amounts to
narrowing of meaning. Both types of narrowing are similar in that they
tend to preserve stable compounds, like {\it meatloaf} or {\it один,
как перст} `lone as a finger'.

There is no symmetry between broadening and narrowing of
meaning. Development of new senses naturally happens all the time
without our really noticing it. But narrowing is typically a result of
competition between words (except for the relatively rare cases where
a word drops out of use because the object it denoted
disappears). Whatever driving forces there were, but {\it hlaf} lost
its generic sense only because it was supplanted by the expanding {\it
bread}, {\it chef} was replaced by the expressive {\it t\^ete} < {\it
testa}, `crock, pot', and {\it перст} by {\it палец} (possibly, also
as an expressive replacement).

This is summarized by Hock and Joseph \cite{HockJoseph} (p.~236):
\begin{quote} 
[...] complete synonymy --- where two phonetically distinct words would
express exactly the same range of meanings --- is highly
disfavored. [...] where other types of linguistic change could give
rise to complete synonymy, we see that languages --- or more
accurately, their speakers --- time and again seek ways to remedy the
situation by differentiating the two words semantically.
\end{quote}
And by Maslov \cite{Maslov}, p.~201:
\begin{quote}
[...] since lexical units of the language are in {\it systemic
  relationships} with each other via semantic fields, synonymic sets,
  and antonymic pairs, it is natural that changes in one element of a
  microsystem entails changes in other related elements.
\end{quote}

One important feature of this process of avoiding excessive synonymy
is that words compete only if their meanings are similar in
scope. That is, a word whose meaning overlaps with that of a
significantly more general word, will not feel the pressure of
competition. As discussed earlier, the language needs (or rather its
speakers need) words of different scope of meaning, so both the more
general and the more specific words retain relevance. This is in a way
similar to the effect reported by Wasow {\it et al}
\cite{WasowAmbiguity} where it was found (both by genetic simulation
and by studying polysemous word use in Brown Corpus) that polysemy
persists if one of the senses is significantly more common than
the other. Despite the fact that this result is related to polysemy
rather than to synonymy, it also can be interpreted as an evidence
that meanings do not interact (compete) if they are sufficiently
different in scope, whether they belong to the same word (polysemy) or
to different words (synonymy).

Summarizing the above, one can say that {\it meanings tend to increase
in scope, unless they collide with other meanings of a similar scope,
while meanings of significantly different scope do not interact}. But
this looks just like a recipe for the development of approximately
Zipfian coverings discussed in the previous section! Indeed, this kind
of evolution could lead to semantic space being covered almost without
gaps and overlaps by each subset of all words of approximately the
same scope. In order to substantiate this idea two numerical models
were developed.

\section{Numerical models}

The models simulate the two basic processes by which word meanings
change in extent: generalization and specialization. They are very
schematic and are not intended to be realistic. We model the semantic
space by the interval $S = [0,1]$ and word meanings by sub-intervals on
it. The evolution of the sub-intervals is governed by the following
algorithms. 

\subsubsection*{Generalization model}
\begin{enumerate}
\item Start with a number $N$ of zero-length intervals $r_i\subset S$
  randomly distributed on~$S$.
\item At each step, grow each interval symmetrically by a small length
  $\delta$, if it is not {\it frozen} (see below). 
\item If two unfrozen intervals intersect, freeze one of them (the
  one to freeze is selected randomly).
\item Go to step 2 if there is more than one unfrozen interval left,
  otherwise stop.
\end{enumerate}

Informally, words in the generalization model have a natural tendency
to extend their meanings, unless this would cause excessive
synonymy. If two expanding words collide, one of them stops
growing. The other one can eventually encompass it completely, but
that is not considered to be ``excessive synonymy'', since by that
time, the growing word is significantly more generic, and words of
different generality do not compete. 

\subsubsection*{Specialization model}
\begin{enumerate}
\item Start with a number $N$ of intervals, whose
  centers are randomly distributed on~$S$ and lengths are uniformly
  distributed on $[0,1]$.
\item For each pair of intervals $r_i$, $r_j$, if they intersect and their lengths $l_i$,
  $l_j$ satisfy $1/\gamma < l_i/l_j < \gamma$, decrease the smaller interval by the length of their intersection.
\item Continue until there is nothing left to change. 
\end{enumerate}

The specialization model simulates avoidance of excessive synonymy
where synonyms compete and one supplants the other in their common
area. Parameter $\gamma$ determines by how much the two words can
differ in extent and still compete.

\begin{figure}[htp]
\caption{Zipf's law generated by specialization and
  generalization models.}\label{fig:modelDistr}
\centering
\input{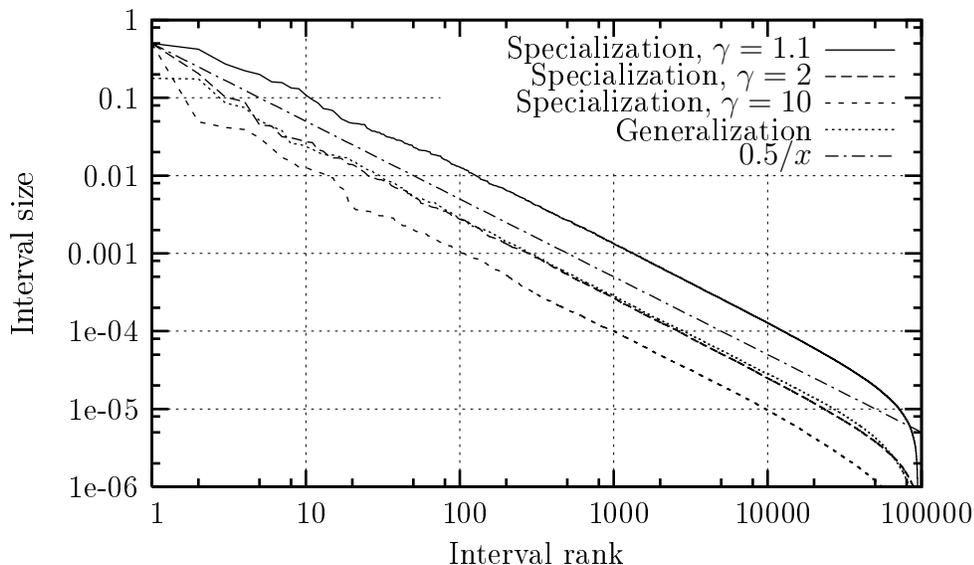}
\end{figure}

Both these models reliably generate interval sets with sizes
distributed by Zipf's law with exponent $B=1$. The generalization
model is parameter-free (except for the number of intervals, which is
not essential as long as it is large enough). The specialization model
is surprisingly robust with respect to its only parameter $\gamma$: we
ran it with $\gamma \in [1.1, 10]$ with the same result --- see
Fig.~\ref{fig:modelDistr}. It is interesting to note that with $\gamma=1.1$,
specialization model even reproduces the low-rank behavior of the actual
rank distributions, but it is not clear whether this is a mere coincidence or
something deeper. 

\begin{figure}[htp]
\caption{The gap of $(k,\rho)$-layer decreases with increasing $k$.}\label{fig:modelGap}
\centering
\input{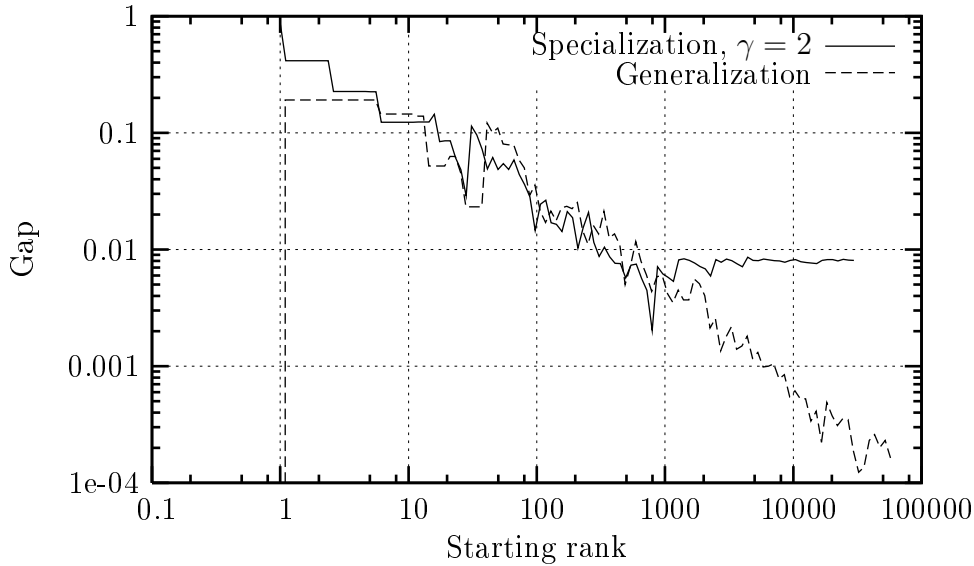}
\end{figure}

Both models also generate interval sizes that approximately satisfy
the definition of Zipfian covering. That is, if we consider the subset
of all intervals between ranks of $k$ and $\rho k$, they should cover
the whole $[0,1]$ interval with no gap and overlap --- for some fixed
$\rho$ and asymptotically in $k\to\infty$.  Fig.~\ref{fig:modelGap} shows the gap,
i.e.~the total measure of that part of $S$ not covered by these
intervals, as a function of the starting rank $k$. Scaling parameter
$\rho$ was chosen so that the sum of interval lengths between ranks
$k$ and $k\rho$ was approximately equal to 1. The fact that the gap indeed
becomes very small demonstrates that the covering is approximately
Zipfian. This effect does not follow from the Zipf's law alone,
because it depends not only on the size distribution, but also on
where the intervals are located on~$S$. On the other hand, Zipf's
distribution does follow from the Zipfianness of the covering.

Of course, these models provide but an extremely crude simulation of
the linguistic processes. However the robustness of the result
suggests that quite possibly they represent a much larger class of
processes that can lead to Zipfian coverings and hence Zipf's
distributions under the same very basic assumptions.

\section{Discussion}

To summarize, we propose the following. 
\begin{enumerate}
\item Word meanings have a tendency to broaden. 
\item On the other hand, there is a tendency to avoid excessive
  synonymy, which counteracts the broadening.
\item Synonymy avoidance does not apply to any two words that differ
  significantly in the extent of their meanings. 
\item As a result of this, word meanings evolve in such a way as to
  develop a multi-layer covering of the semantic space, where each
  layer consists of words of approximately the same broadness of
  meaning, with minimal gap and overlap. 
\item We call arrangements of this sort {\it Zipfian coverings}. It is
  straighforward to show that they possess Zipf's distribution with
  exponent $B=1$. 
\item Since word frequency is likely to be in a direct relationship
  with the broadness of its meaning, Zipf's distribution for
  one of them entails the same distribution for the other. 
\end{enumerate}

This model is rooted in linguistic realities and demonstrates the
evolutionary path for the language to develop Zipf's distribution of
word frequencies. Not only it predicts the power law, but also
explains the specific exponent $B=1$. Even though we argue that
Zipfian coverings are in some sense ``optimal'', we do not need this
optimality to be the driving force, and can in fact do entirely away
with this notion, because the local dynamics of meaning expansion and
synonymy avoidance is sufficient. The ``meaning'' of Zipf's
distribution becomes very clear in this proposal.

The greatest weakness of the model is that it is based upon a rather
vague theory of meaning. The assumption of proportionality of word
frequency to the extent of its meaning is natural (indeed, if one
accepts the view that ``meaning is usage'', it becomes outright
tautological), but it is unverifyable as long as we have no
independent way to measure both quantities or at least compare meaning
extents of different words. On the other hand, comparison of
meaning extent of {\it the same} word at different historical stages
is a less ill-defined notion. See also Appendix~B. Further studies are necessary to clarify
this issue. As one possibility, a direct estimate of word meaning
extent might be obtained on the basis of the Moscow semantic school's Meaning---Text Theory
(e.g. \cite{Melchuk88}, \cite{MelchukCollDeFr}), which
provides a well-developed framework for describing meanings.

The treatment in this work was restricted to the linguistic
domain. However, as is well known, Zipf's law is observed in many
other domains. The mechanism of competitive growth proposed here could
be applicable to some of them. Whenever one has entities that a)
exhibit the tendency to grow, and b) compete only with like-sized
entities, the same mechanism will lead to Zipfian covering of the
territory and consequently to Zipf's distribution of sizes. 

\appendix
\section*{Appendix A: Mandelbrot's model revisited}

Mandelbrot set up to demonstrate that Zipf's law could be
derived from the assumption that the language is optimal in the sense
that it minimizes the average ratio of production cost to information
content. The cost of ``producing'' a word was chosen to be
proportional to the number of letters in it, and information content
was defined to be the Shannon's entropy. It is well known that the
maximum entropy per letter is achieved by random sequences of letters,
just because entropy is a measure of unpredictability, and random
sequences are the most unpredictable. Thus, under these assumptions
the optimal language is the one where each sequence of $n$ letters is
as frequent as any other. But we already know from the analysis of the
random typing model that this does produce the Zipf's distribution. 

Mandelbrot understood well the relationship between his optimality
model and random typing model and remarked in
\cite{Mandelbrot66} that ``these variants are fully equivalent
mathematically, but they appeal to such different intuitions that the
strongest critics of one may be the strongest partisans of
another''. However the optimality model provides a framework that can
be extended beyond this equivalence. 

First of all, let us briefly reproduce the mathematical derivation of
the Zipf's law from the optimality principle. Let $k$ be the frequency
rank of the word $w_k$, let its frequency (normalized so that the sum
of all frequencies is unity) be $p_k$, and the {\it cost} of producing
word $w_k$ be $C_k$. It makes sense to leave the function $C_k$
unspecified for as long as possible. The word's {\it information content}, or entropy,
is related to its frequency $p_k$ as $H_k = -\log_2{p_k}$. The average
cost per word is given by 
\begin{equation}
C = \sum_k{p_kC_k}  \label{defC}
\end{equation}
and the average entropy per word by 
\begin{equation}
H=-\sum_k{p_k\log_2{p_k}}. \label{defH}
\end{equation}
One can now ask what frequency
distribution $\{p_k\}$ satisfying $\sum_k{p_k}=1$ will minimize the
cost ratio $C^* = C/H$.

We can use the standard method of Lagrange multipliers to find the
minimum of $C^*$, given the normalization constraint on $p_k$: 
\begin{equation}
\frac{\partial}{\partial p_k}(C^* + \lambda\sum_j{p_j}) = 0 \label{lagrange}
\end{equation}
Here the value of Langrange multiplier $\lambda$ is to be determined
later so as to normalize the frequencies. Performing the
differentiation in (\ref{lagrange}), we obtain
\begin{equation}
\frac{C_k}{H} + \frac{C}{H^2}(\log_2{p_k} + 1) - \lambda = 0, \forall
k \label{eq2}
\end{equation}
This expresses the frequencies ${p_k}$ given costs ${C_k}$:
\begin{equation}
p_k = \lambda' 2^{-H C_k/C}, \label{solution}
\end{equation}
where we denoted 
\begin{equation}
\lambda' = 2^{\lambda H^2/C-1}.
\end{equation}
Thus, $\lambda'$ is an arbitrary constant that we can use directly to
normalize frequencies. Now, once the cost $C_k$ of each word is known
or assumed, eq.~(\ref{solution}) yields the frequency distribution for the
words. Note though that to obtain a closed-form solution, one also
needs to consistently determine the constants $C$ and $H$ in the RHS
of (\ref{solution}) from their respective definitions (\ref{defC}) and
(\ref{defH}).

Now, it is easy to see from eq.~(\ref{solution}) that a power law for
frequencies could only result from the ansatz
\begin{equation}
C_k = C_0\log_2{k} \label{ansatz1}
\end{equation}
which leads to 
\begin{equation}
p_k = \lambda' k^{-B}, B=H\frac{C_0}{C} \label{powerlaw}
\end{equation}
(note that $C\propto C_0$, so $C_0/C$ doesn't depend on $C_0$). How
could one justify eq.~(\ref{ansatz1})? In Mandelbrot's original
formulation, as we already mentioned, the cost of a word was assumed
to be proportional to its length, and then the only way to get the
logarithmic dependency on the rank, is to assume that the number of
distinct words grows exponentially with length. It is not necessary in
this formulation to postulate that {\it any} combination of letters of
a given length is equally probable, but even this weaker requirement
is not realistic for natural languages, as demonstrated by
Fig.~\ref{wordlenDistr}. 

There is however a much more plausible argument in favor of the
desired ansatz (\ref{ansatz1}), which does not depend on any
assumptions about word length at all. Suppose words are stored in some
kind of an addressable memory. For simplicity, one can imagine a
linear array of memory cells, each containing one word. Then, the cost
of retrieving the word in the $k$-th cell can be assumed to be
proportional to the length of its address, that is to the minimum
number of bits (or neuron firings, say) needed to specify the
address. And this is precisely $\log_2{k}$. Of course, this doesn't
depend on memory being in any real sense ``linear''. 

It's important to note that this is not just a different
justification, because with it the optimality model is no longer
equivalent to the random typing model. Let us now proceed to solving
(\ref{powerlaw}). From the normalization condition for frequencies, we
get
\begin{equation}
p_k = \frac{1}{\zeta(B)}k^{-B} \label{normpowerlaw}
\end{equation}
where $\zeta$ is the Riemann zeta-function $\zeta(s) =
\sum_1^\infty{n^{-s}}$. But this is not the end of the story, since
$B$ is related to $H$ and $C$ via eq.~(\ref{powerlaw}), and they in turn
depend on $B$ via $p_k$. This amounts to an equation for the power law
exponent $B$, which thus is not arbitrary. By substituting
(\ref{normpowerlaw}) back into (\ref{defC}) and (\ref{defH}), we get
\begin{eqnarray}
C &=& \frac{C_0}{\zeta(B)}\sum_1^\infty{k^{-B}\log_2{k}}\\
H &=& \frac{B}{\zeta(B)}\sum_1^\infty{k^{-B}\log_2{(k\zeta(B)^{-1/B})}}
\end{eqnarray}
It is now easy to see that $B=HC_0/C$ can only be satisfied when
$\zeta(B) = 1$, which implies $B\to\infty$. This is not a very
encouraging result, since it means that the minimum cost per unit
information is achieved when there's only one word in use, and both
cost and information vanish. 

This conclusion is borne out by a simple numerical simulation. Recall
that in Section~2, we noted that cost ratio optimization can be
achieved via local dynamics. Namely, if speakers notice that a word's
individual information/cost ratio is below average, they start using
it less, and conversly, if the ratio is favorable, the word's
frequency increases. It is hard to tell {\it a priori} whether this
process would converge to a stationary distribution, so numerical
simulation was performed. The following algorithm implements this dynamics:
\subsubsection*{Cost ratio optimization algorithm}
\begin{enumerate}
\item Initialize an array of $N$ frequencies $p_k$ with random numbers and
  normalize them.
\item Calculate average cost and information per word according to (\ref{defC}),
  (\ref{defH}).
\item For each $k=1,\ldots,N$, calculate cost ratio for the $k$-th
  word as $C^*_k=C_k/H_k=\log_2{k}/\log_2{p_k}$. If it is within the
  interval $[(1-\gamma)C^*, (1+\gamma)C^*]$, where $\gamma$ is a
  parameter, leave $p_k$ unchanged. Otherwise increase $p_k$ by a
  constant factor if cost ratio is above the average or decrease it by the
  same factor if it is below the average.
\item If no frequencies were changed, stop. 
\item Reorder words (i.e.~reassign ranks in the decreasing order of
  frequency), renormalize frequencies and repeat from step 2.
\end{enumerate}
This procedure quickly leads to the state where all frequencies but
one are zero. 

So the ansatz (\ref{ansatz1}) does not eventually lead to the desired
result. It is probably this problem that prompted Mandelbrot to
propose a modification to the Zipf's law. In his own words (\cite{Mandelbrot66}, p. 356), 
\begin{quote}
...it seems worth pointing out that it has {\it not} been obtained by
  ``mere curve fitting'': in attempting to explain the first
  approximation law, $i(r, k)=(1/10)kr^{-1}$, I invariably obtained
  the more general second approximation, and only later did I realize
  that this more general formula was necessary and basically
  sufficient to fit the empirical data.
\end{quote}
It turns out that the degeneracy problem can be avoided by the following
modification of the cost function ansatz:
\begin{equation}
C_k = C_0 \log_2{(k+k_0)} \label{ansatz2}
\end{equation}
It looks rather naturally if we again imagine the linear memory, but
this time with first $k_0$ cells not occupied by useful words. 
Substitution of (\ref{ansatz2}) into (\ref{solution}) yields Zipf--Mandelbrot
law
\begin{equation}
p_k = \frac{1}{\zeta(B, 1+k_0)}(k+k_0)^{-B} \label{normZ-M}
\end{equation}
where $\zeta$ is now the Hurwitz zeta function, $\zeta(s, q) =
\sum_0^\infty{(n+q)^{-s}}$. 

Zipf--Mandelbrot formula has the potential of correctly approximating
not only the power law, but also the initial, low-rank range of the
real frequency distributions, which flatten out at $k<10$ or so. But
remember again that the second part of (\ref{powerlaw}), $B=HC_0/C$,
needs to be satisfied, which means that parameters $k_0$ and $B$ are
not independent. This is rarely, if ever, mentioned in the literature,
while it is a rather important constraint. Substituting
(\ref{normZ-M}) into (\ref{defC}) and (\ref{defH}) and noting that
\begin{equation}
\frac{\partial}{\partial s}\zeta(s, q) =
-\sum_0^\infty{(n+q)^{-s}\ln(n+q)}
\end{equation}
we obtain
\begin{eqnarray}
C &=& -\frac{C_0}{\ln 2}{\displaystyle\frac{\zeta'(B, 1+k_0)}{\zeta(B, 1+k_0)}} \\
H &=& \ln{\zeta(B, 1+k_0)} - {\displaystyle\frac{B}{\ln 2}\frac{\zeta'(B,  1+k_0)}{\zeta(B, 1+k_0)}} \\
B &=& HC_0/C
\end{eqnarray}
where $\zeta'$ is the derivative over the first argument. After 
simple transformations this reduces to 
\begin{equation}
B = B-\frac{\ln\zeta(B, 1+k_0)}{\ln\zeta'(B, 1+k_0)}\ln 2
\end{equation}
that is
\begin{equation}
\zeta(B, 1+k_0) = 1
\end{equation}
When $k_0 \to 0$, $B\to \infty$, as previously. In the oppposite
limit, $k_0\to\infty$, the Zipfian exponent $B$ tends to 1, but
extremely slowly. To see this, let $k_0$ be a large integer. Then, 
\begin{equation}
\zeta(B, 1+k_0) = \zeta(B) - \sum_1^{k_0}{n^{-B}}
\end{equation}
In order to compensate for the infinite growth of the second term as
$k_0\to\infty$, $B$ must tend to 1, where Riemann's zeta function has
a pole. Let $B = 1+\epsilon$, $\epsilon \ll 1$, then
\begin{eqnarray}
\zeta(B) &=& O(1/\epsilon) \\
\sum_1^{k_0}{n^{-B}} &=& O\left(\frac{1}{\epsilon}k_0^{-\epsilon}\right)
\end{eqnarray}
whence $k_0^\epsilon = O(1)$, or $B = 1 + O(1/\ln{k_0})$. 

The relationship between $B$ and $k_0$ can be calculated numerically,
but this would not tell us whether the resulting solution is stable
with respect to the local dynamics described above. Running the local
dynamics model shows that, in contrast to the case $k_0 = 0$, the
model does converge to a stable solution described by (\ref{normZ-M}),
as shown in Fig.~7.

\begin{figure}[htp]
\caption{Zipf--Mandelbrot law with different values of $k_0$. Real
  frequency distribution (not to scale) and Zipf's law are shown for comparison.}\label{fig:7}
\centering
\input{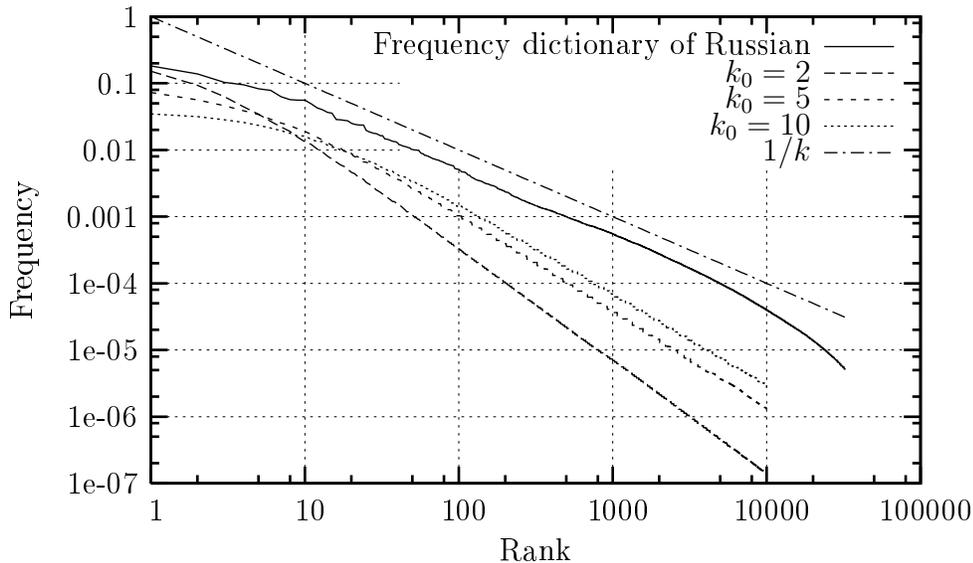}
\end{figure}

However, as is readily seen from the figure, no values of $k_0$ yield
a satisfactory approximation to the actual distribution. For small
$k_0$, the slope is still significantly steeper than $-1$, but for
larger $k_0$, the flattened portion spreads too far. Thus, with
$k_0=10$, the slope is still about $-1.4$, but the power law starts
at about $k=100$, while in the actual distribution it begins after
$k=10$. 

To sum up, Zipf--Mandelbrot law can be obtained from a model
optimizing the information/cost ratio with no assumptions about word
lengths. This model is not equivalent to the random typing model, and
allows the optimum to be achieved via local dynamics, i.e.~in a
causal, rather than teleological manner. However, the distributions
obtained in this way do not provide a reasonable fit to the actual
distributions. In addition, the local dynamics is not convincingly
realistic, as pointed out in Section 2. 

\appendix
\section*{Appendix В: Meaning and frequency}

In this Appendix we'll consider some evidence in favor of the
hypothesis that word frequency is proportional to the extent of its
meaning. Far from being a systematic study, this is rather a
methodological sketch. This study was done in Russian, the author's
native language. In the English text we'll attempt to provide translations
and/or equivalents wherever possible. 

Strictly speaking, one could prove the hypothesis only if an explicit
measure of meaning extent is proposed. However the frequency
hypothesis allows to make some verifiable predictions. Suppose that some
``head'' word $w_0$ has a set of partial synonyms and/or
hyponyms (``specific'' words) $\{w_0^1,\ldots,w_0^n\}$, whose meanings together cover the
meaning of $w_0$ without gaps and overlaps. Then, by definition, 
their total meaning extent is equal to that of $w_0$. In that case,
the frequency hypothesis predicts that the sum total of hyponym
frequencies should be close to the frequency of the head word. 

There's hardly very many such examples in the real language. First,
pure hyponyms are not very common; it is more common for words to have
intersecting meanings, such as with {\it плохой}, `bad, poor', and
{\it худой}, 'skinny; torn, leaky; bad, poor'. Second, only in rare
cases one can state confidently that the hyponyms cover the whole
meaning of the head word. For example, in the domain of fine arts,
{\it натюрморт} `still life', {\it пейзаж} `landscape', and {\it
портрет} `portrait' are pure hyponyms of the word {\it картина}
`picture', but there exist other genres of painting that can't be
accounted for with frequency dictionary, since their names are
phrases, rather than single words ({\it жанровая сцена} 'genre
painting', {\it батальное полотно} 'battle-piece').

Nevertheless, examples of this type do exist. Table \ref{t:tree}
contains frequencies of the head word {\it дерево, деревцо} 'tree;
also dimin.'  and of the specific tree names found in the frequency
dictionary
\cite{FreqDict}. We omitted words denoting primarily the fruit or
bloom of the corresponding tree, such as {\it груша} `pear',
{\it вишня} `sour cherry', {\it рябина} `rowan' или {\it магнолия}
`magnolia'. To count them correctly, one would have to know the
fraction of word instances denoting the tree specifically, and we
don't have this data. 

\begin{table}[htp]
\begin{footnotesize}
\caption{\it Tree.}
\begin{center}
\begin{tabular}{|lr|lr|}
\hline
word & freq./mln & word & freq./mln\\
\hline        	
дерево `tree' & 224.52	& сосна `pine'  & 38.07	\\
деревцо 'tree dimin.' & 8.08	& дуб `oak'   & 27.24	\\
 & & елка `fir'  & 26.57	\\
 & & береза `birch' & 24.36	\\
 & & тополь `poplar' & 17.75	\\
 & & пальма `palm tree' & 16.96	\\
 & & липа `linden'  & 13.89	\\
 & & яблоня `apple tree' & 13.41	\\
 & & ива `willow'   & 7.96	\\
 & & кедр `cedar'  & 7.77	\\
 & & клен `maple'  & 7.53	\\
 & & осина `aspen' & 6.79	\\
 & & лиственница `larch' & 6.00 \\
 & & ель `fir'   & 4.84	\\
 & & орешник `filbert' & 4.84	\\
 & & вяз `elm'   & 3.31	\\
 & & пихта `fir' & 3.24	\\
 & & кипарис `cypress' & 3.18	\\
 & & эвкалипт `eucalyptus' & 2.51	\\
 & & ольха `alder' & 1.96	\\
 & & ясень `ash' & 1.90	\\
 & & ветла `willow' & 1.84	\\
 & & бук `beech'   & 1.78	\\
 & & платан `platan' & 1.71	\\
\hline
sum & {\bf 232.60} & sum & {\bf 246.82}\\
\hline
\end{tabular}\end{center}\end{footnotesize}
\label{t:tree}
\end{table}

From the table one can see that the sum of frequencies of specific tree
names is very close to the frequency of the head word (we'll
consider the ``physicist's error margin'' of 20\% to be
acceptable). Possibly, the word {\it пальма} `palm tree' could be
removed from the list: it is not clear why it turned out to be the
sixth frequent tree in Russian-language texts before {\it липа}
`linden' и {\it яблоня} `apple tree'. However, small changes in the
list will not conceptually affect the result. 

This is just one example of many. Table \ref{t:flower} contains the
frequencies of common flower names. They also sum up very close to the
frequency of the word {\it цветок (цветочек)} 'flower; also
dimin.'. (The word {\it колокольчик} 'small bell; bluebell', frequency 11.08, is
omitted here, since primarily it denotes a bell, and not a flower.)
Possibly, subtracting the frequencies of figurative meanings of words
like {\it роза} `rose', would still improve the result. 

\begin{table}[htp]
\caption{{\it Flower}.}
\begin{footnotesize}

\begin{center}
\begin{tabular}{|lr|lr|}
\hline
word & freq./mln & word & freq./mln\\
\hline        	
цветок `flower'         & 134.85 & роза `rose' & 41.50 \\
цветочек (dimin.)       & 11.87  & мак `poppy' & 27.91 \\
&& тюльпан `tulip'    &12   \\
&& одуванчик `dandellion' &11.32\\
&& сирень `lilac'     &9.92 \\
&& ромашка `daisy'    &8.63 \\
&& лилия   `lily'    &7.65 \\
&& гвоздика  `carnation'  &7.35 \\
&& подсолнух `sunflower'  &5.02 \\
&& черемуха  `bird cherry'  &4.84 \\
&& лютик  `buttercup'     &4.10  \\
&& фиалка  `violet'    &4.22 \\
&& василек  `cornflower'   &3.61 \\
&& ландыш   `lily of the valley'   &2.94 \\
&& хризантема `chrysanthemum' &2.82 \\
&& крокус  `crocus'    &2.26 \\
&& нарцисс `daffodil'    &2.20  \\
&& герань `geranium'     &2.02 \\
&& астра  `aster'     &1.90  \\
&& подснежник `snowdrop' &1.78 \\
&& незабудка `forget-me-not'  &1.65 \\
&& гладиолус `gladiolus'  &1.29 \\
&& орхидея  `orchid'   &1.29 \\
&& пион   `peony'       &1.22    \\
\hline
sum & {\bf 146.72} & sum & {\bf 169.44}\\
\hline
\end{tabular}\end{center}\end{footnotesize}
\label{t:flower}
\end{table}

Names of berries also follow this pattern, see table
\ref{t:berry}. (Here and below, we list in the table captions some
words not found in the dictionary, apparently because their frequency
is less than one per million.) The difference is somewhat greater in
this case, but we should take into account that {\it малина} and {\it
клюква} possess active figurative and idiomatic meanings in Russian
(resp., `a criminal flat' and an approximate equivalent of 'red
herring'). Besides, it is not quite clear whether the cherries {\it
вишня} and {\it черешня} truly belong in this list: first, a
considerable number of instances will refer to corresponding trees,
not fruits, and second, we are not certain whether the designation
{\it ягода} `berry' is appropriate for them. For instance, in the
classical Dahl's dictionary, the entry for {\it cherry} starts with ``A tree
and its frut...'', while the entry for {\it cranberry} or {\it raspberry} starts
with ``A bush and its berry...''. Of course, for the purposes of this
work, it is a matter of lexicography, rather than botany. 

\begin{table}[htp]
\caption{{\it Berry}. Not in dictionary: {\it gooseberry}, {\it cloudberry},
  and{\it bilberry}.}
\center
\begin{footnotesize}\begin{tabular}{|lr|lr|}
\hline
word & freq./mln & word & freq./mln\\
\hline        	
ягода `berry'     & 25.83 & малина `raspberry'    & 7.59	\\
ягодка (dimin.)    & 3.00 & вишня  `sour cherry'    & 6.98	\\
&& земляника `wild strawberry' & 5.69	\\
&& рябина  `rowan berry'  & 3.86	\\
&& смородина `currant' & 3.98	\\
&& клубника `strawberry'  & 3.12	\\
&& клюква  `cranberry'   & 2.94	\\
&& брусника `lingonberry'  & 2.82	\\
&& черника `blueberry'   & 2.69	\\
&& ежевика `blackberry'   & 2.08	\\
&& черешня `cherry'   & 1.47	\\
\hline
sum & {\bf 28.83} & sum & {\bf 43.22}\\
& & without cherries & {\bf 34.77} \\
\hline
\end{tabular}\end{footnotesize}
\label{t:berry}
\end{table}

In all the three examples, we didn't have to face the question of how
to prove that the hyponyms indeed cover the head word's meaning
without overlaps (an object can't be both a gooseberry and a
blueberry) and gaps (each berry has a specific name). However, some
subtleties can already be found here. Thus, if ``в сорок пять баба
ягодка опять'' (a proverb; lit.: ``at 45 a woman is a berry again'')
this ``berry'' is none of the berries we listed. On the other hand,
{\it воровская малина} (`a criminal flat'; lit.: ``thieves'
raspberry'') is not a berry. In this particular case, there is no
doubt that such non-literal usage will not appreciably affect the
results; what's more important, it is possible, at least in principle,
to account for it by studying texts. Below we'll encounter much
greater difficulties, which require systematic and more formal
approaches. 

A somewhat different example is given in table \ref{t:meat},
containing a classification of meat produce, which is pretty chaotic
from a logician's point of view, but quite common in everyday
use. We'll note that although a sausage can contain beef or pork, the
meanings of words {\it колбаса} `sausage' and {\it говядина} `beef' do
not intersect (or intersect negligibly). The same can be said about
other word pairs in the table. For the non-Russian reader, it should
be noted that {\it мясо} does not have many extended meanings of
English {\it meat}, and means practcally nothing beyond 'the flesh of
animals used as food'. But are all the hyponym meanings really
contained within the meaning of the word {\it мясо} `meat'? For
instance, can we say that $\hbox{\it паштет\ \rm `pat\'e'} \subset\hbox{\it
мясо\ \rm `meat'}$ (we will denote the relationships between meanings with
mathematical symbols of subset, intersection, and union $\subset,
\cap, \cup$)? The evidence in favor of this statement is provided by
locutions like {\it Возьми паштет, тебе надо есть больше мяса} ('Take
some pat\'e, you need meat to recover').

\begin{table}[htp]
\caption{{\it Meat}. Not in dictionary: {\it ромштекс} `rump steak', {\it
 шницель} `schnitzel'.}
\center
\begin{footnotesize}\begin{tabular}{|lr|lr|}
\hline
word & freq./mln & word & freq./mln\\
\hline
мясо `meat' & 84.47 & колбаса `sausage, bologna' & 39.48\\
&& котлета `cutlet' & 11.81\\
&& сосиска `sausage' & 9.12\\
&& ветчина `ham' & 6.49\\
&& баранина `(meat of) lamb' & 5.88\\
&& свинина `pork' & 5.82\\
&& бифштекс `steak' & 4.96\\
&& говядина `beef' & 4.22\\
&& фарш `ground meat' & 3.12\\
&& паштет `pat\'e' & 3.06\\
&& телятина `veal' & 2.57\\
&& сарделька `wiener' & 1.78\\
&& отбивная `chop' & 1.47\\
&& котлетка `cutlet (dimin.)' & 1.22\\
\hline        	
sum & {\bf 84.47} & sum & {\bf 101.00}\\
\hline        	
\end{tabular}\end{footnotesize}
\label{t:meat}
\end{table}

So far, we only considered head words from a mid-frequency range
(the most frequent, {\it дерево} `tree' has a rank of 435). But the
supporting data can be found among high-frequency words as well. Table
\ref{t:human} classifies humans by age and gender (the rank of the
word {\it человек} `human, person' is 33; it is counted together with
its plural form, {\it люди}). As an aside, we note the curious fact
that the most frequent words for male and female persons come in
exactly opposite order in terms of age: in the order of decreasing
frequency we have {\it старик} `old man', {\it мальчик} `boy', {\it
парень} `lad, guy', {\it мужчина} `man', but {\it женщина} `woman',
{\it девушка} `young woman', {\it девочка} `young girl', {\it старуха} 'old
woman'. Also, the net frequency of all the male terms (1377) is
practically the same as the net frequency of all the female terms
(1339). Frequency is rather uniformly distributed over age groups as
well. 

\begin{table}[htp]
\caption{{\it Human}.}
\center
\begin{footnotesize}\begin{tabular}{|lr|lr|}
\hline
word & freq./mln & word & freq./mln\\
\hline
человек `human' & 2945.47 & ребенок `child' & 593.50	\\
&& женщина `woman' & 584.32	\\
&& старик `old man'  & 313.64	\\
&& мальчик `boy' & 290.81	\\
&& девушка `young woman' & 286.53	\\
&& парень `lad, guy'  & 258.74	\\
&& мужчина `man' & 252.98	\\
&& девочка `young girl' & 191.04	\\
&& старуха `old woman' & 105.89	\\
&& мальчишка `boy (derog.)'& 92.55	\\
&& девица `girl; virgin'  & 59.86	\\
&& девчонка  `young girl (derog.)'      & 58.95		\\
&& юноша  `young man'  & 58.09	\\
&& старушка  `old woman (dimin.)'      & 52.21		\\
&& старичок `old man (dimin.)' & 40.95	\\
&& пацан `boy (dial., colloq.)' & 24.91	\\
&& младенец `baby'       & 27.18		\\
&& паренек `boy, dimin. of {\it lad}'  & 21.73	\\
&& парнишка `boy, dimin. of {\it lad}'  & 19.95		\\
&& дитя `child'    & 17.02	\\
&& мальчонок `boy (dimin.)' & 3.00	\\
\hline        	
sum & {\bf 2945.47} & sum & {\bf 3353.85} \\
&& without neut. terms & {\bf 2716.15} \\
\hline        	
\end{tabular}\end{footnotesize}
\label{t:human}
\end{table}

There are new difficulties in this case: obviously, there are
significant intersections between the meanings of some hyponyms. This
is mostly because $$\hbox{\it мальчик, девочка \rm `boy, girl'}\subset
(\hbox{\it ребенок \rm `child'}\cup\hbox{\it дитя \rm `child'}\cup\hbox{\it
младенец \rm `baby'})$$ (a boy or a girl is almost necessarily a child or a
baby) \footnote{Of course, there are exceptions here, too. Compare a
quote from abovementioned Viktor Konetsky: {\it A fiftyish grocery
store saleswoman is universally called ``девушка'' (girl), even though
she has five children. And I once heard older female road workers
going for lunch say: ``Let's go, girls!''} Such a {\it girl} is not a
{\it child}.}. Indeed, the net frequency of the words {\it ребенок},
{\it дитя}, {\it младенец} `child, baby' is 637.7, and the net
frequency of the words {\it мальчик}, {\it девочка}, {\it мальчишка},
{\it девчонка}, {\it пацан}, {\it паренек}, {\it парнишка}, {\it
мальчонок} `boy, girl' is 702.94, which is pretty close. So we can
subtract the net frequency of the neutral terms from the sum of
frequencies, which makes the net frequency of the rest of hyponyms
very close to the frequency of the head word {\it человек} `human'.

The frequency hypothesis works with words of relatively low frequency
as well: see tables \ref{t:fish} ({\it рыба} `fish') and \ref{t:fence}
({\it забор} `fence').

\begin{table}[htp]
\caption{{\it Fish}. Not in dictionary: {\it красноперка `rudd', салака
`sprat', палтус `halibut', ставрида `scad', нотатения, тунец `tuna',
кефаль `mullet', налим `burbot', плотва `roach', севрюга `sturgeon',
пескарь `gudgeon', мурена `moray', омуль `omul'.}}
\center
\begin{footnotesize}\begin{tabular}{|lr|lr|}
\hline
word & freq./mln & word & freq./mln\\
\hline
рыба `fish' & 120.03& сазан	`sazan' & 16.47	\\
рыбка (dimin.)& 20.02& карась	`crucian' & 14.63	\\
&& акула `shark'        &10.77 \\
&& селедка `herring'      & 9.61 \\
&& карп    `carp'      & 9.24 \\
&& щука   `pike'       & 9.06 \\
&& сом    `catfish'       & 8.20  \\
&& скат   `ray'       & 6.98 \\
&& судак  `pike perch'       & 6.06 \\
&& лещ    `bream'       & 5.51 \\
&& форель `trout'       & 4.53 \\
&& окунь  `perch'       & 4.41 \\
&& вобла  `vobla'       & 2.94 \\
&& камбала `flounder'      & 2.88 \\
&& угорь  `eel'       & 2.82 \\
&& лосось  `salmon'      & 2.57 \\
&& треска  `cod'      & 2.14 \\
&& сельдь  `herring'      & 2.08 \\
&& хек      `hake'     & 2.02 \\
&& семга   `salmon'      & 1.78 \\
&& осетр   `sturgeon'      & 1.59 \\
&& ерш     `ruff'      & 1.59 \\
&& сардина `sardine'      & 1.53 \\
&& стерлядь `sterlet'     & 1.47 \\
&& скумбрия `mackerel'     & 1.22 \\
&& белуга  `beluga'      & 1.10  \\
&& горбуша `salmon'      & 1.10  \\
\hline        	
sum & {\bf 140.05} & sum & {\bf 134.43}\\
\hline        	
\end{tabular}\end{footnotesize}
\label{t:fish}
\end{table}

\begin{table}[htp]
\caption{{\it Fence}. Not in dictionary: {\it палисад}.}
\center
\begin{footnotesize}\begin{tabular}{|lr|lr|}
\hline
word & freq./mln & word & freq./mln\\
\hline
забор `fence'& 66.72& ограда `fence' & 25.83 \\
&& изгородь `fence, hedge'     &10.59 \\
&& плетень  `wicker fence'     & 9.61 \\
&& частокол `stake fence'    & 5.39 \\
&& штакетник `picket fence'    & 2.57 \\
&& загородка `fence'    & 2.20  \\
&& тын  `paling'         & 1.96 \\
\hline        	
sum & {\bf 66.72} & sum & {\bf 58.15}\\
\hline        	
\end{tabular}\end{footnotesize}
\label{t:fence}
\end{table}

\begin{table}[htp]
\caption{{\it Old}. Not in dictionary: {\it закоснелый, заматорелый,
	затасканный, зачерствелый, истасканный, подержанный,
	полинялый, поседелый, потрепанный, старобытный}.} 
\center
\begin{footnotesize}\begin{tabular}{|lr|lr|}
\hline
word & freq./mln & word & freq./mln\\
\hline
старый `old'	&	528.25	&	древний `ancient'	&	75.60	\\
	&		&	пожилой `elderly'	&	63.17	\\
	&		&	седой `grey-haired'	&	62.99	\\
	&		&	старинный `antique'	&	53.07	\\
	&		&	давний `bygone'	&	34.71	\\
	&		&	бородатый `bearded; old (of jokes)'	&	18.67	\\
	&		&	немолодой `not young'	&	16.34	\\
	&		&	многолетний `longstanding'	&	11.51	\\
	&		&	старомодный `old-fashioned'	&	11.51	\\
	&		&	престарелый `very old (of people)'	&	10.04	\\
	&		&	ветхий `shabby, decrepit'	&	9.67	\\
	&		&	вековой	`age-old' &	6.86	\\
	&		&	извечный `primeval'	&	6.67	\\
	&		&	отсталый `outdated, retrograde'	&	5.94	\\
	&		&	дряхлый	`decrepit'&	5.82	\\
	&		&	устарелый `outmoded, outdated'	&	5.39	\\
	&		&	ископаемый `fossilized'	&	5.20	\\
	&		&	поношенный `worn, shabby'	&	4.77	\\
	&		&	допотопный `antediluvian'	&	4.16	\\
	&		&	давнишний `bygone'	&	3.55	\\
	&		&	застарелый `inveterate'	&	3.37	\\
	&		&	многовековой `centuries-old'	&	3.37	\\
	&		&	исконный `original'	&	3.06	\\
	&		&	заскорузлый `calloused, backward'	&	2.69	\\
	&		&	закоренелый `inveterate, ingrained'	&	1.96	\\
	&		&	истертый `worn'	&	1.71	\\
	&		&	отживший `obsolete'	&	1.65	\\
	&		&	архаический `archaic'	&	1.35	\\
	&		&	стародавний `ancient'	&	1.35	\\
	&		&	обветшалый `shabby, decrepit'	&	1.29	\\
	&		&	архаичный `archaic'	&	1.04	\\\hline        	
sum & {\bf 528.25} & sum & {\bf 438.48}\\
\hline        	
\end{tabular}\end{footnotesize}
\label{t:old}
\end{table}

\begin{table}[htp]
\caption{{\it Red}. Not in dictionary: {\it карминный, рдяный, червленый}.}
\center
\begin{footnotesize}\begin{tabular}{|lr|lr|}
\hline
word & freq./mln & word & freq./mln\\
\hline
красный `red'	&	316.64	&	рыжий `red-haired; rust-colored'	&	89.8	\\
	&		&	розовый	`rosy, pink' &	77.98	\\
	&		&	алый `scarlet'	&	32.99	\\
	&		&	кровавый `bloody'	&	32.93	\\
	&		&	багровый `crimson'	&	22.16	\\
	&		&	румяный	`ruddy' &	17.2	\\
	&		&	малиновый `crimson'	&	14.02	\\
	&		&	пунцовый `crimson'	&	3.55	\\
	&		&	бордовый `vinous'	&	2.82	\\
	&		&	багряный `crimson (arch., poet.)'	&	2.63	\\
	&		&	коралловый `coral'	&	2.57	\\
	&		&	морковный `carrot (adj.)'	&	2.57	\\
	&		&	рубиновый `ruby (adj.)'	&	2.2	\\
	&		&	пурпурный `purple'	&	1.84	\\
	&		&	свекольный `beet (adj.)'	&	1.04	\\
\hline        	
sum & {\bf 316.64} & sum & {\bf 306.3}\\
\hline        	
\end{tabular}\end{footnotesize}
\label{t:red}
\end{table}

Let us now consider other parts of speech. Two simple examples with
adjectives can be found in tables \ref{t:old} ({\it старый} `old') and
\ref{t:red} ({\it красный} `red'). A more complicated example is given
by the word {\it большой} `big, large' shown in table \ref{t:big}. The
net frequency of hyponyms significantly (by a quarter) exceeds the
frequency of the head word. This is as expected, since some of the
hyponyms' meanings definitely intersect: thus, {\it огромный} and {\it
громадный} are as close to exact synonyms as it gets (cf. Eng. {\it
huge} and {\it enormous}). However there's a possibility for a deeper
and more interesting analysis here.

Consider locutions \ref{ili1.1}--\ref{ili3.4}. 
\begin{eqnarray}
&\hbox{Is this a raspberry or a strawberry?} \label{ili1.1}\\
&\hbox{*Is this a strawberry or a berry?} \label{ili1.2}\\
&\hbox{Is this a boy or a girl?} \label{ili2.1}\\
&\hbox{Is this a boy or a man?} \label{ili2.2}\\
&\hbox{*Is this a boy or a child?} \label{ili2.3}\\
&\hbox{*Is this a boy or a person?} \label{ili2.4}\\
&\hbox{Do you want pork or pat\'e?} \label{ili3.1}\\
&\hbox{(?)Купить свинину или мясо? '$\approx$Do you want pork or meat?'} \label{ili3.2}\\
&\hbox{(?)Купить паштет или мясо? '$\approx$Do you want pat\'e or meat?'} \label{ili3.3}\\
&\hbox{*Купить говядину или мясо? '$\approx$Do you want beef or meat?'} \label{ili3.4}
\end{eqnarray}

Everything is clear with items \ref{ili1.1}--\ref{ili3.1}:
non-intersecting specific words can occur in alternative constructions
with each other, but not with the head words. Locutions
\ref{ili3.2}, \ref{ili3.3} are possible only if {\it мясо} `meat' is used in
constrained, specialized (sub)meanings, existing in the vernacular:
$\hbox{(\it мясо \rm `meat')}^2=\hbox{\it говядина \rm `beef'}$,
$\hbox{(\it мясо \rm `meat')}^3=\hbox{\it сырое мясо \rm `raw meat'}$
(this is proved by the fact that \ref{ili3.4} is not possible). These
example, therefore, also involve non-intersecting (non-overlapping)
meanings. As a first approximation, we will consider this as a
criterion of meaning overlap: if two words can participate in an
alternative construction of this type, their meanings do not overlap.

To apply this criterion to hyponyms of the word {\it большой} 'big,
large', consider examples \ref{ili4.1}, \ref{ili4.2}. Although the
semantic difference between them is intuitively obvious, it is not
easy to explicate it. There are objects that are both long and wide,
as well as objects that are both long ang huge, --- and still the
first example is perfectly valid, while the second one is
impossible. But keeping with the methodological principles of this
work, we will not attempt to formulate the difference in semantic
terms. On the contrary, we take acceptability of a locution as a
linguistic datum, and on this basis draw conclusions about word
semantics. That is, we will {\it define} two words non-overlapping in
their meanings, if they can can participate in an alternative
construction of the type \ref{ili1.1}--\ref{ili4.2}.

\begin{eqnarray}
&\hbox{Is it wide or long?} \label{ili4.1}\\
&\hbox{*Is it long or huge?} \label{ili4.2}
\end{eqnarray}

Now, accepting the above criterion for non-overlapping meanings, we
can select a subset of hyponyms from table \ref{t:big}, which do not
overlap and mean roughly `big/large in a certain dimension or
trait'. Almost all remaining hyponyms are in fact emphatic or
superlative terms: `very big/large, regardless of dimension or trait'
(only two, {\it немалый} `not small' and {\it изрядный} 'fairly
large', are hard to classify). It is easy to make sure that the first
group consists of virtually non-overlapping adjectives. Admittedly, in
the lower part of the table, the criterion becomes less clear-cut:
thus, the question in example \ref{ili4.3} is somewhat awkward;
however it is meaningful and understandable, in contrast to
\ref{ili4.2}. Of course, there is still some overlap in the meanings;
after all, we're dealing with a living language. But it is small
enough so that any further corrections will not change the result in
any significant way (and may still improve it). 

\begin{equation}
\vcenter{
 \hbox{--- The king's palace is big!}
 \hbox{--- Is it spacious or grandiose?} 
}
\label{ili4.3}
\end{equation}

In the 5th column of table \ref{t:big} we sum up the frequencies of
the hyponyms that are specifying the trait or dimension. The net
frequency is very close to the frequency of the head word. 

\begin{table}[htp]
\caption{{\it Big/large}.}
\center
\begin{footnotesize}\begin{tabular}{|lr|lr|l|c|}
\hline
word & freq./mln & word & freq./mln & trait & emphasis\\
\hline
большой & 1630.96 & высокий `tall, high' & 310.34 & height & -\\
\ \ \ `big, large'  && огромный `huge' & 298.95 & - & +\\
&& великий `great (significant)' & 247.90 & significance & +\\
&& длинный `long (space)' & 244.05 & length & -\\
&& широкий `wide' & 187.31 & width & -\\
&& толстый `thick' & 176.12 & diameter; thickness & -\\
&& крупный `large-scale, coarse' & 151.74 & all dimensions & -\\
&& глубокий `deep' & 135.58 & depth & -\\
&& долгий `long (time)'& 132.52 & time & -\\
&& значительный `significant'& 60.17 & significance & -\\
&& гигантский `giant'& 42.24 & - & +\\
&& громадный `tremendous' & 40.77 & - & +\\
&& длительный `prolonged'& 35.56 & time & -\\
&& просторный `spacious'& 28.03 & space & -\\
&& обширный `vast' & 26.20 & extent & -\\
&& немалый `not small'& 22.83 & - & -\\
&& грандиозный `grandiose'& 18.24 & impression, intent & +\\
&& внушительный `impressive'& 13.34 & impression & -\\
&& колоссальный `colossal'& 9.79 & - & +\\
&& громоздкий `bulky'& 9.73 & all dims.; maneuverability & -\\
&& изрядный `fairly large'& 8.75 & - & -\\
&& исполинский `gigantic'& 6.37 & - & +\\
&& масштабный `large-scale'& 4.16 & intent; influence & -\\
&& непомерный `exorbitant' & 3.98 & - & +\\
&& объемный `bulky'& 3.43 & bulk, volume & -\\
&& объемистый `voluminous'& 3 & volume, bulk & -\\
&& большущий `big (superl.)' & 2.14 & - & +\\
&& протяженный `lengthy'& 1.47 & length & - \\
\hline        	
sum & {\bf 1630.96} & sum & 2048.59 & {\bf 1788.89} & 670.38 \\
\hline        	
\end{tabular}\end{footnotesize}
\label{t:big}
\end{table}

\begin{table}[htp]
\caption{{\it Small}.}
\center
\begin{footnotesize}\begin{tabular}{|lr|lr|l|c|}
\hline
word & freq./mln & word & freq./mln & trait & emphasis\\
\hline
маленький & 411.52 & короткий `short in length' &202.55 & length& - \\
\ \ \ `small, little' & & тонкий `thin' & 144.58 & thickness & - \\
небольшой & 180.08 & мелкий `shallow; fine' & 125.05 &depth; all dims. & - \\
\ \ \ `not large'  &  & узкий `narrow'& 105.47 & width & - \\
малый  & 108.71  & низкий `low; short in height'& 78.23 & height & - \\
\ \ \ `lesser; too small'  &  & тесный `tight'& 33.18 & spaciousness & - \\
 &  & крохотный `tiny' & 28.4 & - & + \\
 &  & крошечный `tiny' & 24.67 & - & + \\
 &  & незначительный `insignificant' & 20.69 & significance & - \\
 &  & ничтожный `very insignificant'& 19.71 & significance & + \\
 &  & невеликий `not great' & 13.04 & significance & - \\
 &  & миниатюрный `miniature' & 5.26 & - & + \\
 &  & неглубокий `not deep' & 4.77 & depth & - \\
 &  & неширокий `not wide' & 3.86 & width & - \\
 &  & малюсенький `small (superl.)' & 3.61 & - & + \\
 &  & мизерный `paltry' & 3.61 & - & + \\
 &  & микроскопический `microscopic' & 3.06 & - & + \\
 &  & махонький `wee' & 2.02 & - & + \\
 &  & недлинный `not long' & 1.78 & length & - \\
\hline        	
sum & {\bf 700.31} & sum & 823.54 & {\bf 752.91} & 90.33 \\
\hline        	
\end{tabular}\end{footnotesize}
\label{t:small}
\end{table}

The word {\it маленький} `small, little' (table \ref{t:small}) is very
similar. However we face a new complication here: the main concept is
expressed by three words, rather than one: {\it маленький}, {\it
небольшой} and, possibly, {\it малый}. It is somewhat similar to the
distinction between {\it small} and {\it little} in English. Consider
the first two adjectives. Both are direct and stilistically neutral
antonyms to {\it большой} `big, large'. However their meanings are
distinct. For example, they are not interchangeable in the common
phrases like {\it маленький мальчик} `little boy' and {\it небольшое
количество} `small amount': {\it *небольшой мальчик} and {\it
*маленькое количество} are not normative (while the adjective {\it
большой} `big, large' can modify both nouns). But even when both
adjectives are admissible, they mean different things. Thus, {\it
маленькая мышка} `$\approx$a little mouse' means `small compared to
the speaker, as all mice', or, less probably, `a young mouse', but
{\it небольшая мышка} `$\approx$a small mouse' means `small compared
to other mice, less than usual mouse size'. Even when this distinction
is not applicable, there still can be a quantitative difference, as in
example \ref{ili5}.

\begin{equation}
\vcenter{
 \hbox{--- Этот кусок слишком большой. `This piece is too big.'}
 \hbox{--- Отрезать тебе небольшой или маленький? '$\approx$Do you want a
 smaller one or a small one?'} 
}
\label{ili5}
\end{equation}

As a result, we consider the words {\it маленький} и {\it небольшой}
to have almost non-overlapping meanings. As for the adjective {\it
малый}, in its long form it is used only in compound toponyms and
scientific nomenclature (cf. {\it Lesser Antilles}). But in its short
form, it has a common and distinctive meaning of `too small to fit',
not covered by adjectives {\it маленький} and {\it небольшой}. Indeed,
if {\it туфли малы} `$\approx$shoes are too small', this doesn't
necessarily mean that the shoes are small, they still can be size
10. But they are necessarily narrow, short, or tight. This is why the
adjective {\it малый} is also placed in table \ref{t:small} as a head
word, and not as a hyponym.

This argument is based on intuitive judgement about acceptability of
certain expressions, which is not a very solid foundation
(cf. \cite{WasowIntuition}). To improve it, one would have to
formulate strict criteria of intersection and inclusion for meanings,
and then demonstrate that they are satisfied. This is generally beyond the scope
of the present essay, but one example of a completely objectivised
approach is given below for the word {\it плохой} `bad'.

Verbs provide some good examples as well. See tables \ref{t:say} ({\it
сказать} `say') and \ref{t:think} ({\it думать} `think') that do not
require any comments.

\begin{table}[htp]
\caption{{\it Say}.}
\center
\begin{footnotesize}
\begin{tabular}{|lr|lr|}
\hline
word & freq./mln & word & freq./mln\\
\hline
сказать `say'	&	3535.97	& спросить `ask (a question)'	&	934.32	\\
&& ответить `answer'	&	503.46	\\
&& рассказать `tell'	&	248.58	\\
&& произнести `pronounce'	&	178.98	\\
&& крикнуть `shout'	&	155.97	\\
&& попросить `ask (for a favor)'	&	154.62	\\
&& сообщить `inform'	&	148.80	\\
&& приказать `command'	&	107.18	\\
&& велеть `order'	&	95.67	\\
&& заявить `state'	&	86.61	\\
&& воскликнуть `exclaim'	&	81.66	\\
&& проговорить `utter'	&	78.35	\\
&& возразить `object'	&	69.66	\\
&& предупредить `caution'	&	51.23	\\
&& пробормотать `mutter'	&	49.52	\\
&& прошептать `whisper'	&	39.79	\\
&& пообещать `promise'	&	33.42	\\
&& возмутиться `say indignantly'	&	27.61	\\
&& осведомиться `inquire'	&	26.32	\\
&& буркнуть `growl'	&	25.52	\\
&& шепнуть `whisper'	&	24.79	\\
&& пошутить `joke'	&	24.06	\\
&& поздороваться `greet'	&	22.34	\\
&& выразиться `curse'	&	22.28	\\
&& попрощаться `say goodbye'	&	20.51	\\
&& скомандовать `command'	&	19.59	\\
&& проворчать `growl'	&	18.79	\\
&& рявкнуть `bark out'	&	17.14	\\
&& выговорить `utter'	&	16.22	\\
&& прокричать `shout'	&	12.67	\\
&& высказаться `express'	&	12.12	\\
&& провозгласить `announce'	&	11.94	\\
&& гаркнуть `bawl'	&	10.04	\\
&& молвить `say (arch., poet.)'	&	9.67	\\
&& промолвить `say (arch., poet.)'	&	6.18	\\
&& брякнуть `blurt'	&	6.18	\\
&& пролепетать `babble'	&	5.08	\\
&& промямлить `mumble'	&	4.47	\\
&& съязвить `say sarcastically'	&	3.67	\\
&& вопросить `inquire'	&	2.88	\\
&& вякнуть `blather'	&	2.51	\\
&& предостеречь `warn'	&	2.45	\\
\hline
sum & {\bf 3535.97} & sum & {\bf 3372.85}\\
\hline
\end{tabular}
\end{footnotesize}
\label{t:say}
\end{table}

\begin{table}[htp]
\caption{\it Think.}
\center
\begin{footnotesize}
\begin{tabular}{|lr|lr|}
\hline
word & freq./mln & word & freq./mln\\
\hline
думать `think'	&	936.40	&	считать `reckon'	&	396.22	\\
	&		&	мечтать `dream'	&	83.61	\\
	&		&	полагать `believe'	&	73.45	\\
	&		&	предполагать `presume'	&	50.56	\\
	&		&	рассуждать `reason'	&	38.20	\\
	&		&	соображать `consider'	&	36.36	\\
	&		&	размышлять `reflect on'	&	29.75	\\
	&		&	воображать `imagine'	&	20.69	\\
	&		&	мыслить `concieve'	&	19.47	\\
	&		&	раздумывать `ponder'	&	16.53	\\
	&		&	прикидывать `reckon'	&	11.57	\\
	&		&	обдумывать `think over'	&	11.14	\\
	&		&	вникать `fathom'	&	7.53	\\
	&		&	помышлять `dream of'	&	3.80	\\
	&		&	замышлять `scheme'	&	2.75	\\
	&		&	мнить `imagine'	&	2.33	\\
	&		&	вдумываться `ponder'	&	1.47	\\
	&		&	кумекать `think (low colloq.)'	&	1.16	\\
\hline
sum & {\bf 936.40} & sum & {\bf 806.59}\\
\hline
\end{tabular}
\end{footnotesize}
\label{t:think}
\end{table}

\begin{figure}[htp]
\caption{{\it Rise} and {\it grow} (cf. table \ref{t:grow}).}\label{fig:grow}
\centering
\includegraphics{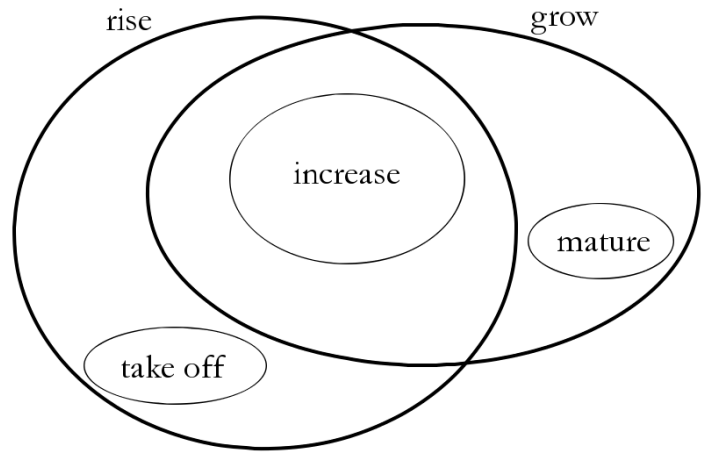}%\fig{img/figGrow}
\end{figure}

In two other verbs we encounter a complication of a new type: see
tables \ref{t:grow} ({\it подниматься/расти} 'rise/grow') and
\ref{t:cry} ({\it кричать/плакать} 'shout/cry'). The words {\it
подниматься} `rise, ascend' and {\it расти} `grow, increase' have some
common sub-meanings, such as {\it увеличиваться} 'increase in quantity
or size' as well as distinct ones, such as {\it взлетать} 'soar, take
off' and {\it расширяться} `widen, spread' respectively. For example,
a temperature can both rise and grow (in Russian ``температура
растет'' is much more common than in English ``temperature grows''),
these expressions being quite synonymous and meaning the increase in
temperature. On the other hand, an elevator can only rise, while a
child can only grow (a child can rise up on the toes, but this is a
completely different meaning, of course). Apparently, in every or
almost every context, the verb {\it увеличиваться} `increase' can be
replaced with either {\it подниматься} `rise' or {\it расти} `grow'
(this is a statement about Russian verbs, not their approximate
equivalents in English), which means that its meaning is a subset of
the intersection of their meanings --- see Fig. \ref{fig:grow}.

It turns out that the net frequencies of hyponyms match the head word
frequencies in both columns of table \ref{t:grow}. This would even
allow to quantify the degree of commonality between the meanings of
the two head words. Exactly the same behavior can be observed with
words {\it кричать} `shout' and {\it плакать} `cry'.

\begin{table}[htp]
\caption{{\it Rise} and {\it grow}. Translations are very approximate.}
\center
\begin{footnotesize}\begin{tabular}{|lr|lr|}
\hline
word & freq./mln & word & freq./mln\\
\hline

подниматься `rise' & {\bf 102.41} & расти `grow' & {\bf 71.74}\\
\hline
				
увеличиваться `increase' & 21.24 & увеличиваться `increase' & 21.24\\
вырастать `grow' & 13.04 & вырастать `grow' & 13.04\\
возрастать `grow' & 12.12 & возрастать `grow' & 12.12\\
прибывать `rise, swell' & 12.12 & прибывать `grow, swell' & 12.12\\
взлетать `soar up, take off' & 14.38 && \\
взбираться `climb' & 8.81 && \\
& & расширяться `spread, widen' & 8.32\\
всплывать `rise to the surface' & 6.92 && \\
вздыматься `heave' & 5.51 && \\
подрастать `grow' & 4.77 & подрастать `grow' & 4.77\\
восходить `rise, ascend' & 4.04 && \\
всходить `rise, ascend' & 3.98 && \\
возноситься `rise, tower' & 1.71 & возноситься `rise, tower' & 1.71\\
& & взрослеть `mature' & 2.94\\
& & шириться `expand, widen' & 2.69\\
& & совершенствоваться `improve' & 2.69\\
взвиваться `soar up, be hoisted' & 1.78 && \\
& & умножаться `multiply' & 1.16\\
\hline
sum & {\bf 108.64} & sum & {\bf 82.8}\\
\hline
\end{tabular}\end{footnotesize}
\label{t:grow}
\end{table}

\begin{table}[htp]
\caption{{\it Shout} and {\it cry}. Translations are very approximate.}
\center
\begin{footnotesize}\begin{tabular}{|lr|lr|}
\hline
word & freq./mln & word & freq./mln\\
\hline

кричать `shout'	&	{\bf 220.36}	&	плакать `cry, weep'	&	{\bf 120.71}	\\
\hline

орать `yell'	&	67.64	&		&		\\
шуметь `make noise'	&	44.62	&		&		\\
реветь `roar; cry'	&	26.99	&	реветь `roar; cry'	&	26.99	\\
	&		&	рыдать `sob'	&	18.30	\\
выть `wail'	&	17.51	&	выть `wail'	&	17.51	\\
визжать `shriek'	&	16.04	&	визжать `shriek'	&	16.04	\\
вопить `bawl'	&	15.98	&		&		\\
	&		&	всхлипывать `sob'	&	12.06	\\
надрываться `bawl'	&	6.86	&		&		\\
	&		&	скулить	`whine' &	4.84	\\
галдеть	`clamor' &	4.41	&		&		\\
	&		&	пищать `squeak'	&	4.28	\\
верещать `chirp, squeal'	&	3.67	&		&		\\
скандалить `brawl'	&	3.67	&		&		\\
голосить `wail'	&	3.24	&		&		\\
	&		&	хныкать `whimper'	&	2.20	\\
горланить `bawl'	&	1.84	&		&		\\
гомонить `shout'	&	1.35	&		&		\\
\hline
sum & {\bf 213.82} & sum & {\bf 102.22}\\
\hline
\end{tabular}\end{footnotesize}
\label{t:cry}
\end{table}

Finally, consider two more adjectives, {\it хороший} `good' (table
\ref{t:good}) and {\it плохой} `bad, poor' (table \ref{t:bad}). Synonyms (or
rather hyponyms) were collected from dictionaries. The former word
doesn't cause any difficulties: the net frequency of hyponyms
corresponds well with the head word frequency. However, with the
adjective {\it плохой} `bad' the situation is quite different. Note
first of all that the four most frequent synonyms offered by the
dictionaries ({\it худой} `skinny; leaky; bad', {\it низкий} 'low,
short; base, mean', {\it дешевый} `cheap, worthless', {\it жалкий}
`pitiful; wretched') are not included in the table, because each of
them has a primary meaning that does not directly imply
badness. Something or somebody can be cheap and good, skinny and good,
etc. But even without them, the net frequency of hyponyms is
significantly over the head word frequency.

Notice though that the hyponyms can be roughly classified into two
categories: those denoting more of an objective quality of an object,
like {\it скверный} (cf. Eng. {\it poor} in its senses unrelated to
pitying and lack of wealth), and those denoting more of a subjective
feeling towards the object, like {\it мерзкий} `loathsome, vile'. The
head word itself falls more in the former category. To demonstrate
this, consider the expression {\it плохой вор} `a bad thief'. Its
meaning is `one who is not good at the art of stealing', in contrast
to {\it мерзкий вор} `vile thief' $=$ `one whom I loath because he
steals'. Hence, only the frequencies of the hyponyms from the first
category (denoting quality) should sum up to the frequency of the head
word.

\begin{table}[htp]
\caption{\it Good.}
\center
\begin{footnotesize}
\begin{tabular}{|lr|lr|}
\hline
word & freq./mln & word & freq./mln\\
\hline
хороший `good'	&	853.71	&	добрый `good, kind'	&	201.38	\\
	&		&	прекрасный `splendid, excellent'	&	143.17	\\
	&		&	приятный `nice'	&	74.31	\\
	&		&	блестящий `brilliant'	&	61.33	\\
	&		&	замечательный `remarkable'	&	60.84	\\
	&		&	благородный `noble'	&	57.66	\\
	&		&	отличный `excellent'	&	42.24	\\
	&		&	славный `glorious, nice'	&	38.44	\\
	&		&	великолепный `magnificent'	&	34.95	\\
	&		&	чудесный `wonderful'	&	34.46	\\
	&		&	роскошный `splendid'	&	27.91	\\
	&		&	неплохой `not bad'	&	26.38	\\
	&		&	чудный `wonderful'	&	13.71	\\
	&		&	превосходный `excellent'	&	13.47	\\
	&		&	прелестный `lovely, delightful'	&	12.24	\\
	&		&	дивный `charming'	&	9.79	\\
	&		&	благой `good'	&	8.88	\\
	&		&	безупречный `impeccable'	&	8.63	\\
	&		&	образцовый `exemplary'	&	8.57	\\
	&		&	годный `suitable, valid'	&	7.96	\\
	&		&	путный `worthwhile'	&	7.77	\\
	&		&	отменный `excellent'	&	7.35	\\
	&		&	изумительный `marvellous'	&	6.79	\\
	&		&	восхитительный `adorable'	&	6.55	\\
	&		&	пригодный `suitable'	&	6.49	\\
	&		&	добросовестный `conscientious'	&	4.53	\\
	&		&	удовлетворительный `satisfactory'	&	3.86	\\
	&		&	доброкачественный `of good quality'	&	3.31	\\
	&		&	благоустроенный `well-furnished'	&	2.08	\\
	&		&	похвальный `laudable'	&	1.96	\\
	&		&	бесподобный `incomparable'	&	1.78	\\
\hline
sum & {\bf 853.71} & sum & {\bf 938.79}\\
\hline
\end{tabular}
\end{footnotesize}
\label{t:good}
\end{table}

\begin{table}[htp]
\caption{{\it Bad}. Some translations are very approximate.}
\center
\begin{footnotesize}
\begin{tabular}{|lr|lr|r|c|}
\hline
word & freq./mln & word & freq./mln & weight & quality? \\
\hline
плохой `bad, poor' & 102.22 & дурной `bad, mean'	&	40.40	& 0.911 & + \\
&&противный `repugnant'	&	28.34	& -0.0584 &  \\
&&отвратительный `disgusting'	&	21.85	& -0.439  &  \\
&&нехороший `not good'	&	20.14	& 0.914 & + \\
&&мерзкий `vile'	&	13.22	& -1.946 &  \\
&&скверный `bad, poor'	&	13.16	& 0.896 & + \\
&&гнусный `abominable'	&	12.73	& -3.160 &  \\
&&поганый `foul'	&	11.51	& -0.330 &  \\
&&паршивый `nasty'	&	10.16	& -0.407 &  \\
&&кошмарный `nightmarish'	&	9.30	& -0.180 &  \\
&&негативный `negative'	&	7.10	& -0.183 &  \\
&&неважный `rather bad'	&	6.00	& 1.200 & + \\
&&омерзительный `disgusting'	&	6.00	& -0.432 &  \\
&&гадкий `repulsive; nasty'	&	5.33	& -0.490 &  \\
&&хреновый `bad, poor (colloq.)'	&	5.14	& 2.358 & + \\
&&никчемный `worthless'	&	5.08	& 0.144 & + \\
&&негодный `worthless'	&	4.10	& 0.157 & + \\
&&дрянной `rotten, trashy'	&	3.92	& -0.110 &  \\
&&никудышный `worthless'	&	3.37	& 3.095 & + \\
&&захудалый `run-down'	&	2.57	& 0.347 & + \\
&&неприглядный `unsightly'	&	2.39	& --- &  \\
&&незавидный `unenviable'	&	1.90	& -0.161 &  \\
&&дерьмовый `shitty'	&	1.90	& 0.161 & + \\
&&фиговый `bad, poor (colloq.)'	&	1.78	& 0.545 & + \\
&&неудовлетворительный `unsatisfactory'	&	1.65	& -0.077 &  \\
&&паскудный `foul, filthy'	&	1.59	& -0.203 &  \\
&&отвратный `disgusting'	&	1.41	& -0.165 &  \\
&&грошовый `dirt-cheap'	&	1.35	& -0.172 &  \\
&&бросовый `worthless, trashy'	&	1.35	& --- &  \\
&&пакостный `foul, mean'	&	1.35	& -0.234 &  \\
&&одиозный `odious'	&	1.35	& -0.122 &  \\
&&сволочной `mean, vile'	&	1.04	& -0.318 &	 \\
&&аховый `rotten'	&	0	& -0.109 &  \\
&&дефектный `defective'	&	0	& -0.179 & \\
&&завалящий `worthless'	&	0	& -0.078 &  \\
&&мерзостный `disgusting'	&	0	& -0.270 &  \\
&&мерзопакостный `disgusting'	&	0	& -0.302 &  \\
&&низкопробный `low-grade'	&	0	& -0.406 &  \\
&&отталкивающий `revolting'	&	0	& -0.198 &  \\
\hline
sum & {\bf 102.22} & & 248.48 & & {\bf 103.64}\\
\hline
\end{tabular}
\end{footnotesize}
\label{t:bad}
\end{table}

But it is quite difficult to actually classify the words into these
two categories. The ``subjective'' words tend to evolve towards
emphatic terms, and further migrate to the ``objective'' group or
close to it. So we need a method that would allow to perform
classification without relying on dubious judgements based on the
linguistic intuition. To this end, notice that there exist three
classes of nouns by their compatibility with the adjectives from table
\ref{t:bad}. Neutral nouns, like {\it погода} `weather' can be equally
easy found in noun phrases with both {\it скверный} `bad, poor' and {\it
мерзкий} '$\approx$disgusting'. However the nouns carrying distinct
negative connotation, such as {\it предатель} `traitor' are well
compatible with {\it мерзкий} '$\approx$disgusting', but not with {\it
скверный} `bad, poor'. On the contrary, nouns with distinct positive
connotation have the opposite preference: cf. {\it скверный
поэт} `bad poet' and ?*{\it мерзкий поэт} `disgusting poet'. It is
possible to find out which of the adjectives in table \ref{t:bad}
tend to apply preferentially to positive or negative nouns, by using an
Internet search engine. 

We considered eight test nouns: negative {\it
гадость} '$\approx$filth', {\it дрянь} '$\approx$trash', {\it
  предатель} `traitor', {\it предательство} `treason' and positive
{\it здоровье} `health', {\it врач} `doctor', {\it поэт} `poet', {\it
  актер} `actor'. They were initially selected for maximum contrast in
their compatibility with adjectives {\it скверный} and {\it
  мерзкий}. Then we used Russian-specific search engine Yandex
(http://www.yandex.ru) to determine the frequencies of noun phrases
constructed from each of the adjectives with each of the nouns. 

It should be noted here that search engines can't be directly used as
replacements for a frequency dictionary. First, they typically report
the number of ``pages'' and ``sites'', but not the number of word
instances. Meanwhile, web pages can be of very different size, and may
contain multiple instances of a word or search phrase. Second, search
engines trim the results to exclude ``similar pages'' and avoid
duplicates, i.e. texts available in multiple copies or from multiple
addresses. It's not clear whether this is correct behavior from the
point of view of calculating frequencies. Finally, the corpus with
which search engines work, the whole of the Web, is by no means
well-balanced according to the criteria of frequency dictionary
compilers. So the results from search engines can't be directly
compared with the data from frequency dictionaries. But for our
purposes we need only relative figures, and we are interested in their
qualitative behavior only. The effect we are looking for, if it
exists, should be robust enough to withstand the inevitable
distortion.

\begin{table}[htp]
\caption{Compatibility of the hyponyms of {\it плохой} `bad, poor'
  with test nouns on the Web (``the number of pages'').}
\center
\begin{footnotesize}
\begin{tabular}{|l|r|r|r|r|r|r|r|r|}
\hline
 & дрянь & гадость & предатель & предательство & здоровье & врач & актер & поэт\\
\hline
дурной & 0 & 1 & 0 & 0 & 81 & 66 & 187 & 172\\
противный & 140 & 333 & 45 & 0 & 0 & 61 & 35 & 31\\
отвратительный & 305 & 5187 & 0 & 112 & 71 & 45 & 211 & 43\\
нехороший & 15 & 38 & 11 & 19 & 24 & 282 & 4 & 11\\
мерзкий & 627 & 1354 & 849 & 316 & 7 & 42 & 30 & 16\\
скверный & 4 & 3 & 15 & 3 & 232 & 54 & 250 & 137\\
гнусный & 156 & 62 & 1380 & 1934 & 2 & 1 & 0 & 3\\
поганый & 183 & 27 & 97 & 33 & 87 & 14 & 17 & 32\\
паршивый & 493 & 32 & 156 & 12 & 27 & 8 & 314 & 38\\
кошмарный & 26 & 39 & 0 & 4 & 0 & 7 & 8 & 1\\
негативный & 2 & 0 & 0 & 0 & 0 & 0 & 0 & 0\\
неважный & 0 & 0 & 0 & 0 & 1589 & 22 & 62 & 141\\
омерзительный & 149 & 183 & 10 & 80 & 0 & 5 & 9 & 0\\
гадкий & 132 & 257 & 140 & 28 & 12 & 3 & 15 & 3\\
хреновый & 1 & 0 & 0 & 0 & 226 & 166 & 431 & 381\\
никчемный & 58? & 0 & 7 & 0 & 33 & 23 & 38 & 81\\
негодный & 63 & 0 & 6 & 0 & 108 & 39 & 32 & 58\\
дрянной & 114 & 2 & 0 & 0 & 6 & 1 & 18 & 62\\
никудышный & 13 & 0 & 0 & 0 & 136 & 146 & 989 & 409\\
захудалый & 0 & 1 & 0 & 0 & 0 & 8 & 22 & 150\\
неприглядный & 0 & 0 & 0 & 0 & 0 & 0 & 0 & 0\\
незавидный & 0 & 0 & 0 & 0 & 39 & 0 & 0 & 0\\
дерьмовый & 8 & 1 & 51 & 1 & 13 & 14 & 89 & 70\\
фиговый & 0 & 0 & 0 & 0 & 11 & 33 & 53 & 167\\
неудовлетворительный & 0 & 0 & 0 & 0 & 212 & 0 & 0 & 0\\
паскудный & 4 & 4 & 18 & 4 & 0 & 1 & 0 & 0\\
отвратный & 28 & 92 & 0 & 0 & 2 & 10 & 12 & 1\\
грошовый & 0 & 0 & 0 & 0 & 0 & 0 & 6 & 0\\
бросовый & 0 & 0 & 0 & 0 & 0 & 0 & 0 & 0\\
пакостный & 34 & 22 & 4 & 3 & 0 & 0 & 0 & 0\\
одиозный & 0 & 0 & 7 & 0 & 0 & 0 & 28 & 8\\
сволочной & 99 & 21 & 18 & 0 & 16 & 2 & 0 & 0\\
аховый & 0 & 0 & 0 & 0 & 3 & 0 & 3 & 21\\
дефектный & 0 & 0 & 0 & 0 & 3 & 0 & 0 & 0\\
завалящий & 1 & 0 & 0 & 0 & 0 & 26 & 2 & 0\\
мерзостный & 41 & 36 & 36 & 0 & 0 & 1 & 0 & 0\\
мерзопакостный & 96 & 84 & 4 & 2 & 2 & 4 & 0 & 1\\
низкопробный & 173 & 29 & 0 & 2 & 0 & 0 & 6 & 1\\
отталкивающий & 0 & 3 & 19 & 0 & 0 & 0 & 1 & 0\\
\hline
Eigenvector & -0.300 & -0.110 & -0.418 & -0.377 & 0.206 &	0.355 & 0.416 &	0.489 \\
\hline
\end{tabular}
\end{footnotesize}
\label{t:bad.correl}
\end{table}

The frequencies of noun phrases constructed from each of the
adjectives $a_i$ with each of the test nouns $n_j$ form a matrix ${N_{ij}}$
presented in table \ref{t:bad.correl}. One can readily see that the
rows ``мерзкий'' and ``скверный'' clearly separate the test nouns into
two groups preferentially compatible with one or the other. Many other
rows of the table (e.g. ``гнусный'' and ``неважный'') behave in the
same way. But there are rows that do not, and that is precisely the
reason to consider multiple test words. Thus the adjective {\it
негодный} '$\approx$worthless' is well compatible with all the
positive test nouns, but also with the negative test noun {\it дрянь}
'$\approx$trash'. The adjectives {\it неприглядный} `unsightly' and
{\it бросовый} `worthless', as it turns out, are not compatible with
any of them, so they are excluded from further analysis. Their low
frequency can't appreciably change the result anyway. 

To recap, we want to classify the rows of table \ref{t:bad.correl} by
whether each row is more similar to the row ``скверный'' (quality of
the object) or to the row ``мерзкий'' (speaker's attitude towards the
object). This can be done via statistical procedure known as principal
component analysis or method of empirical eigenfunctions.

First, each row of table \ref{t:bad.correl} was normalized by
subtracting the average and dividing by the standard deviation. This
makes the rows ``мерзкий'' and ``скверный'' roughly opposite to each
other: positive on positive test nouns and negative on negative ones,
or vice versa. Then, correlation matrix of the table's columns was
calculated (size $8\times 8$) and its first eigenvector
$n^1_j$. Finally, the eigenvector's scalar products with $i$-th row of
the table yields the weight of the corresponding adjective $a^1_i =
\sum_j{n^1_jN_{ij}}$.

Mathematically, the result of this procedure is that the product
$a^1_in^1_j$ provides the best (in terms of mean square) approximation
of this kind to the matrix $N_{ij}$. In other words, each row of the
normalized table \ref{t:bad.correl} is approximately proportional to
the pattern row $n^1_j$ multiplied by the weight $a^1_i$. The pattern
row is given at the bottom of table \ref{t:bad.correl}. As expected,
it correctly classifies test nouns as positive and negative. This
means that they actually behave in opposite ways relative to the
adjectives of interest. Now we can classify all the adjectives with
positive weights $a^1_i > 0$ as proper hyponyms of the word {\it
плохой} `bad, poor'. The weights are shown in table \ref{t:bad}
(in an arbitrarily normalization). The table shows that the net
frequency of these proper hyponyms is very close to the frequency of
the head word. 

So it can be seen that the frequency hypothesis is confirmed here as
well, and this conclusion is not based on any intuitive judgement
about word semantics. 

We conclude with a brief discussion of some encountered
counterexamples. In contrast to the words {\it дерево} `tree', {\it
цветок} `flower', {\it ягода} `berry', and {\it рыба} `fish', the
words {\it животное} `animal' and, to a lesser extent, {\it птица}
`bird' are significantly less frequent than predicted by the net
frequency of their hyponyms. The reason is probably that some of the
most frequent animal and bird names have very wide connotations, far
beyond the notion of 'this or that animal/bird'; e.g. {\it осел}
`donkey, ass' and {\it орел} `eagle' (apparently, a much less loaded
word in English than in Russian, where it readily stands for power,
grandeur, nobility, both straight and ironic). It is not surpirising
then, that the frequency of such words is much greater than had they
denoted strictly the corresponding animals. (See also the discussion
of the words {\it собака} `dog' and {\it лошадь} `horse' in Section
2). Among tree and flower names, only a small number are like that,
and to a much smaller degree, e.g. {\it дуб} `oak' (its Russian
figurative meaning as `a dumb, insensitive person' doesn't seem to
have a counterpart in English) and {\it роза} `rose' (which doesn't
have any fixed dictionary senses other than the flower, but has an
established tradition of metaphoric usage). It is possible, at least
in principle, to quantify the last statement by analyzing the actual
word usage, and then counterexamples could turn into confirming
evidence.

Interesting counterexamples are provided by words {\it страна}
`country, state', {\it город} `city, town', {\it река} `river, creek',
and {\it озеро} `lake'. The net frequency of the nouns {\it страна}
`country, state', {\it государство} `state, nation', {\it республика}
`republic', and {\it королевство} `kingdom' is 705.39 per mln. The net
frequency of all the countries of the world found in the dictionary
(except the former Soviet republics) is 1206.05, which is about 70\%
too much. However the first word in the list, {\it Россия} `Russia',
is four times as frequent as the number two ({\it Германия}
`Germany'). Its frequency is 358.88 per mln and is responsible for
most of the discrepancy. Of course, Russia for Russian speakers is
much more than just another country. Most of the rest of the
discrepancy can be attributed to the fact that the word {\it Америка}
`America' denotes two continents and a part of world, in addition to
the country.

A very similar is the situation with the word {\it город} 'city,
town'. Its frequency is 630.59 per mln, while the net frequency of all
city names we could find in the dictionary is 1087.18. But here again,
{\it Москва} `Moscow' (frequency 420.89, 5--6 times more than the next
city name) is responsible for the whole discrepancy. ``Москва...
как много в этом звуке''\footnote{``Moscow... how much the sound embraces'',
from Pushkin's {\it Eugene Onegin}}. 

On the other hand, the net frequency of all the river names in the dictionary is
somewhat less than the frequency of the word {\it река} `river'
(187.61 vs. 199.36), and that despite the fact that {\it дон}, {\it
  Урал}, and {\it Амур} are not just river names (a Spanish nobleman
title, the Ural mountains, and `Cupid; love affair'
respectively). This same effect is much more pronounced with the word
{\it озеро} `lake': its frequency is 74.496 while the net frequency of
all the lake names in the dictionary is only 21.72. Most probably,
this is because only five lake names made it to the dictionary: {\it
Байкал} `Baikal', {\it Ладога} `Ladoga', {\it Онега} `Onega', {\it
Виктория} `Victoria' (some instances are, probably, personal
names), {\it Иссык-Куль} `Issyk-Kul'. Most lake names either fall
below the 1 per mln threshold, or are homonymous with common names or
adjectives. The same is true to a lesser degree for river names. 

To summarize, we demonstrated on several examples that the hypothesis
of word frequency being proportional to the extent of its meaning is
supported by available data, while counterexamples are few and tend to
have plausible explanations. Of course, a much more thorough and
systematic investigation is in order until the hypothesis can be
considered proven. We only sketched some promising approaches to such
an investigation. But it also should be noted that the examples
considered span a wide range of word frequencies, include all three
main parts of speech, and involve very common words, not specially
hand-picked ones.

\bibliographystyle{unsrt}
\bibliography{all}

\begin{thebibliography}{10}

\bibitem{Zipf49}
G.~K. Zipf.
\newblock {\em Human behavior and the principle of least effort}.
\newblock Addison-Wesley, 1949.

\bibitem{FreqDict}
S.~Sharoff.
\newblock The frequency dictionary for {R}ussian.

\bibitem{Sharoff2002}
Serge Sharoff.
\newblock Meaning as use: exploitation of aligned corpora for the contrastive
  study of lexical semantics.
\newblock In {\em {Proc. of Language Resources and Evaluation Conference
  (LREC02)}}, Las Palmas, Spain, May 2002.

\bibitem{FerrerVar05}
R.~{Ferrer~i~Cancho}.
\newblock The variation of {Z}ipf's law in human language.
\newblock {\em Eur.~Phys.~J.~B}, 44:249--257, 2005.

\bibitem{Mandelbrot66}
B.~Mandelbrot.
\newblock Information theory and psycholinguistics: A theory of word
  frequencies.
\newblock In P.~F. Lazarsfield and N.~W. Henry, editors, {\em Readings in
  mathematical social sciences}, pages 350--368. MIT Press, Cambridge, 1966.

\bibitem{Li92}
Wentian Li.
\newblock Random texts exhibit {Z}ipf's-law-like word frequency distribution.
\newblock {\em IEEE Transactions on Information theory}, 38(6):1842--1845,
  1992.

\bibitem{Shan48}
C.~E. Shannon.
\newblock A mathematical theory of communication.
\newblock {\em The Bell System Technical Journal}, 27(3):379--423, 1948.

\bibitem{Mandelbrot82}
B.~Mandelbrot.
\newblock {\em The Fractal geometry of Nature}.
\newblock Freeman, New York, 1982.

\bibitem{Simon55}
H.~A. Simon.
\newblock On a class of skew distribution functions.
\newblock {\em Biometrica}, 42:425--440, 1955.

\bibitem{Simon57}
H.~A. Simon.
\newblock {\em Models of Man}, chapter 6: On a class of skew distribution
  functions.
\newblock John Wiley and Sons, New York, 1957.

\bibitem{Yule1925}
G.~Yule.
\newblock A mathematical theory of evolution based on the conclusions of {D}r.
  {J}.~{C}.~{W}illis.
\newblock {\em F.R.S. Philosophical Transactions of the Royal Society of London
  (Series B)}, 213:21--87, 1925.

\bibitem{Guiraud68}
Pierre Guiraud.
\newblock The semic matrices of meaning.
\newblock {\em Social Science Information}, 7(2):131--139, 1968.

\bibitem{Mandelbrot53}
B.~Mandelbrot.
\newblock An informational theory of the statistical structure of languages.
\newblock In W.~Jackson, editor, {\em Communication Theory}, pages 486--502.
  Butterworth, Woburn, MA, 1953.

\bibitem{ArapovShrejder78}
M.~V. Arapov and Yu.~A. Shrejder.
\newblock Zakon cipfa i princip dissimmetrii sistem.
\newblock In {\em Semiotics and Informatics}, volume~10, pages 74--95. VINITI,
  Moscow, 1978.
\newblock In Russian.

\bibitem{BalasubraNaranan2002}
V.~K. Balasubrahmanyan and S.~Naranan.
\newblock Algorithmic information, complexity and {Z}ipf's law.
\newblock {\em Glottometrics}, (4):1--26, 2002.

\bibitem{FerrerLeastEffort05}
R.~{Ferrer~i~Cancho}.
\newblock Decoding least effort and scaling in signal frequency distributions.
\newblock {\em Physica~A}, 345:275--284, 2005.

\bibitem{FerrerZipfExponent05}
R.~{Ferrer~i~Cancho}.
\newblock Hidden communication aspects in the exponent of zipf's law.
\newblock {\em Glottometrics}, 11:98--119, 2005.

\bibitem{ManinGlotto}
Yu.~I. Manin.
\newblock K probleme rannikh stadij rechi i soznaniya (filogenez).
\newblock In E.~P. Velikhov and A.~V. Chernavsky, editors, {\em Intellectual
  Processes and their Modeling}, pages 154--178. Nauka, Moscow, 1987.
\newblock In Russian.

\bibitem{ManinRKS}
Yu.~I. Manin.
\newblock Expanding constructive universe.
\newblock In A.~P. Ershov and D.~Knuth, editors, {\em Algorithms in modern
  mathematics and computer science : proceedings, Urgench, Uzbek SSR, September
  16-22, 1979}, volume 122 of {\em Lecture notes in computer science},
  Berlin--New York, 1981. Springer-Verlag.

\bibitem{KonetskySL}
Viktor Konetsky.
\newblock {\em Vcherashnie zaboty. {S}olenyj Led.}, page 508.
\newblock Izvestiya, Moscow, 1980.
\newblock In Russian.

\bibitem{AkmajianEtAl95}
A.~Akmajian, R.~M. Harnish, R.~A. Demers, and F.~K. Farmer.
\newblock {\em Linguistics. An introduction to language and communication}.
\newblock MIT Press, Cambridge, 1995.

\bibitem{Carrol94}
D.~W. Carroll.
\newblock {\em Psychology of language}, chapter 9: Conversational interaction,
  pages 242--248.
\newblock Brooks/Cole Publishing Company, Pacific Grove, 1994.

\bibitem{HockJoseph}
H.~H. Hock and B.~D. Joseph.
\newblock {\em Language History, Language Change, and Language Relationship}.
\newblock Mouton de Gruyter, Berlin, New York, 1996.

\bibitem{Maslov}
Yu.~S. Maslov.
\newblock {\em Vvedenie v yazykoznanie}.
\newblock Vysshaya shkola, Moscow, 2nd edition, 1987.
\newblock In Russian.

\bibitem{Traugott}
Elizabeth~C. Traugott and Richard~B. Dasher.
\newblock {\em Regularity in Semantic Change}.
\newblock Cambridge Univ. Press, Cambridge, 2005.

\bibitem{CambridgeEncLang}
David Crystal.
\newblock {\em The Cambridge Encyclopedia of Language}.
\newblock Cambridge Univ. Press, Cambridge, 2nd edition, 2003.

\bibitem{WasowAmbiguity}
Thomas Wasow, Amy Perfors, and David Beaver.
\newblock The puzzle of ambiguity.
\newblock In O.~Orgun and P.~Sells, editors, {\em Morphology and The Web of
  Grammar: Essays in Memory of Steven G. Lapointe}. CSLI Publications,
  Stanford, 2005.

\bibitem{Melchuk88}
I.~Mel'\v{c}uk.
\newblock Semantic description of lexical units in an {E}xplanatory
  {C}ombinatorial {D}ictionary: Basic principles and heuristic criteria.
\newblock {\em International Journal of Lexicography}, 1(3):165--188, 1988.

\bibitem{MelchukCollDeFr}
I.~Mel'\v{c}uk.
\newblock Vers une linguistique sens-texte. {L}e\c{c}on inaugurale.
\newblock Paris: Coll\`ege de France., 1997.

\bibitem{WasowIntuition}
Thomas Wasow and Jennifer Arnold.
\newblock Intuitions in linguistic argumentation.
\newblock {\em Lingua}, 115(11):1481--1496, 2005.

\end{thebibliography}

\end{document}